\documentclass[runningheads]{llncs}

\usepackage{microtype}
\usepackage{graphicx}
\graphicspath{./img/}
\usepackage{subfigure} 

\usepackage{booktabs} 

\usepackage{hyperref}

\usepackage{amsmath}
\usepackage{amssymb}
\usepackage{mathtools}
\usepackage{multirow}
\usepackage{bm}
\usepackage{xcolor}
\usepackage[misc]{ifsym}
\usepackage[normalem]{ulem}

\newcommand{\spd}{\operatorname{SPD}}

\newcommand{\vertices}{\mathcal{V}}
\newcommand{\edges}{\mathcal{E}}
\newcommand{\neighs}{\mathcal{N}}
\newcommand{\RR}{\mathbb R}

\newcommand{\similar}{\operatorname{sim}}
\DeclareMathOperator*{\argmax}{arg\,max}

\newcommand{\insertDimensions}{
\begin{figure}%
\centering
\includegraphics[width=0.5\linewidth]{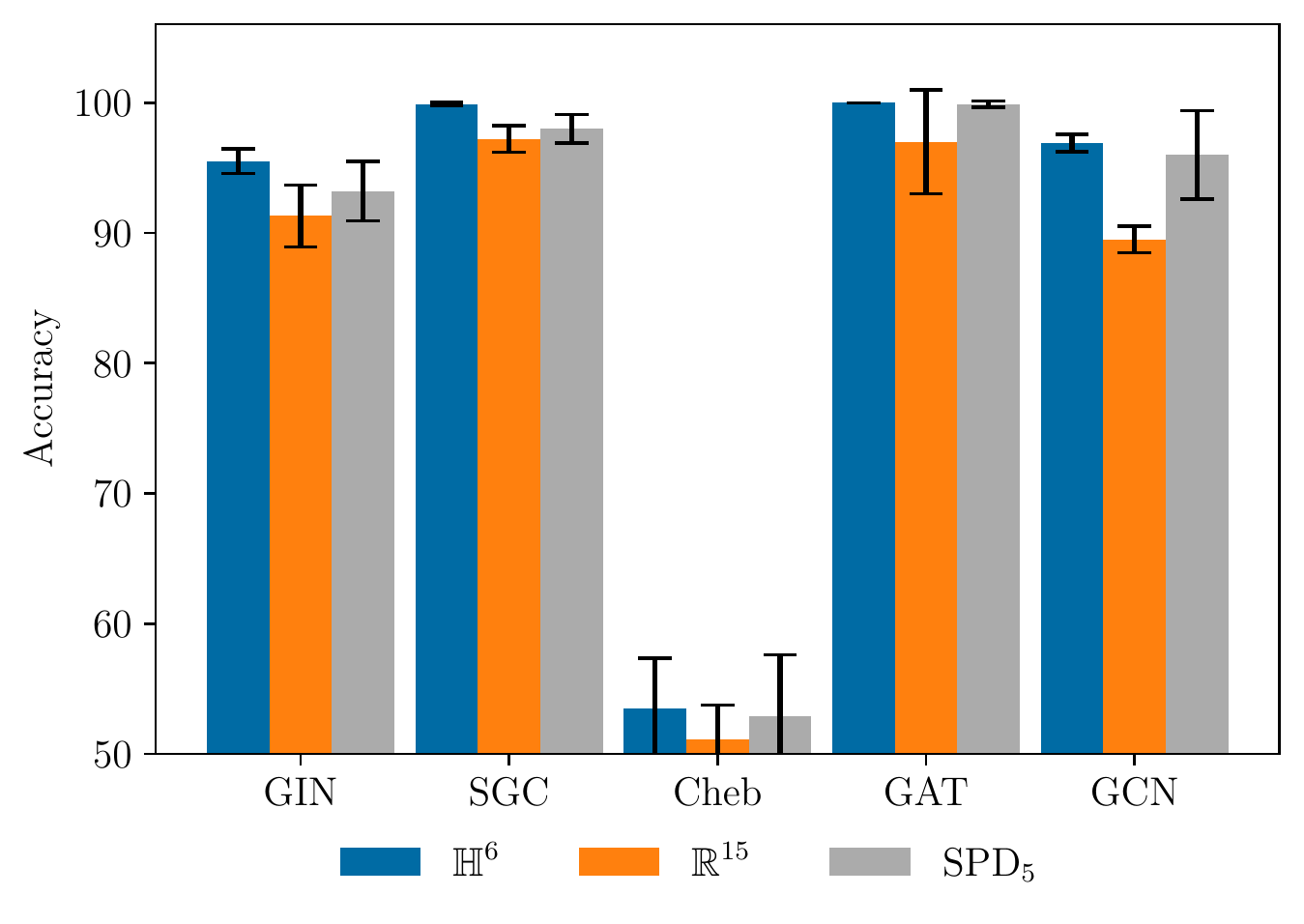}
\caption{Comparison of 6d and 15d spaces on Disease. $\spd_5$ has 15 dimensions. }
\label{fig:6d_15d}
\end{figure}
}

\newcommand{\insertDatasetsA}{
\begin{figure*}[htbp]

\centering
{%
    \subfigure[Disease ($\delta=0.0$)]{%
    \includegraphics[width=0.3\linewidth]{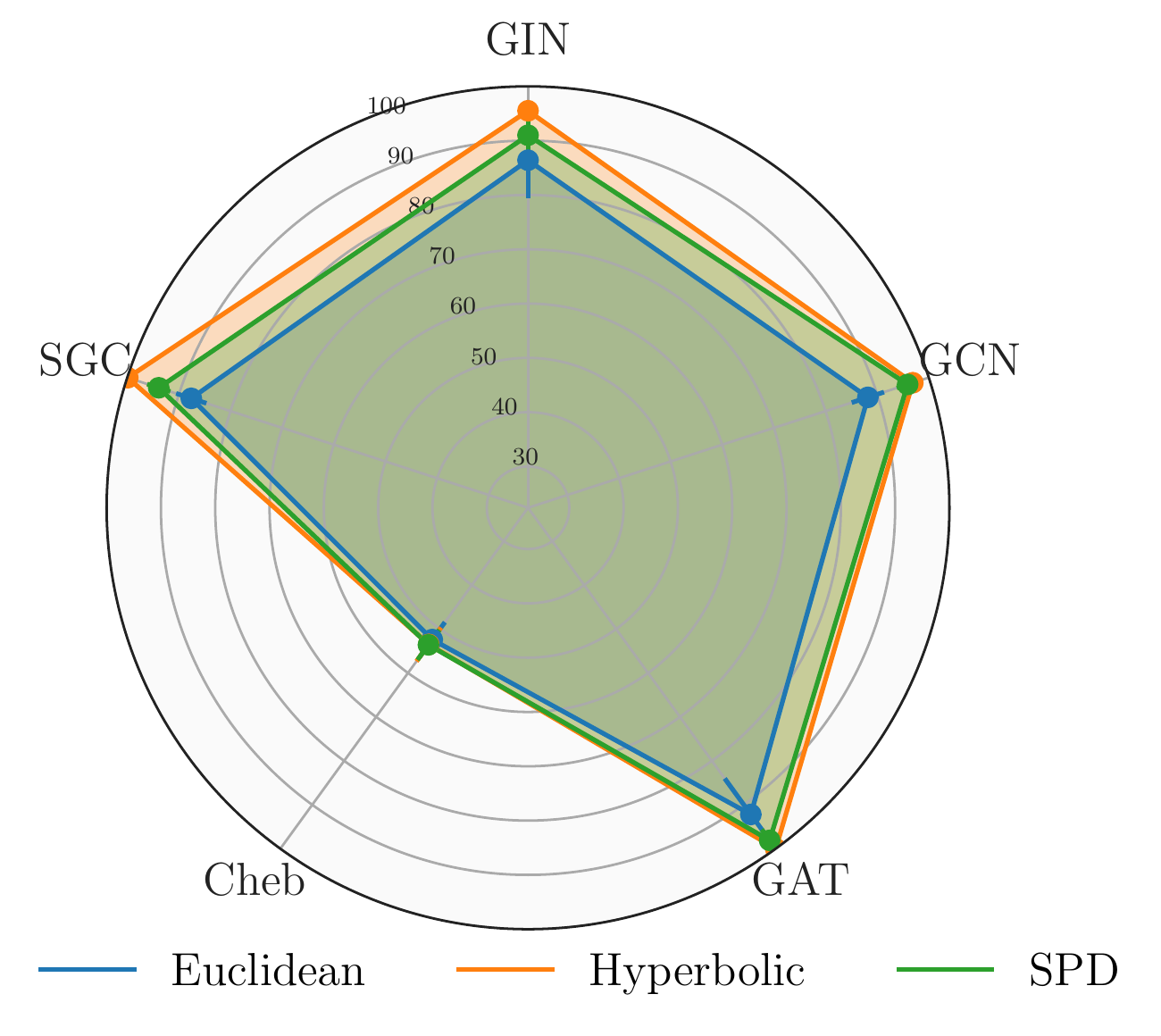}
    } 
    \subfigure[Airport ($\delta=1.0$)]{%
    \includegraphics[width=0.3\textwidth]{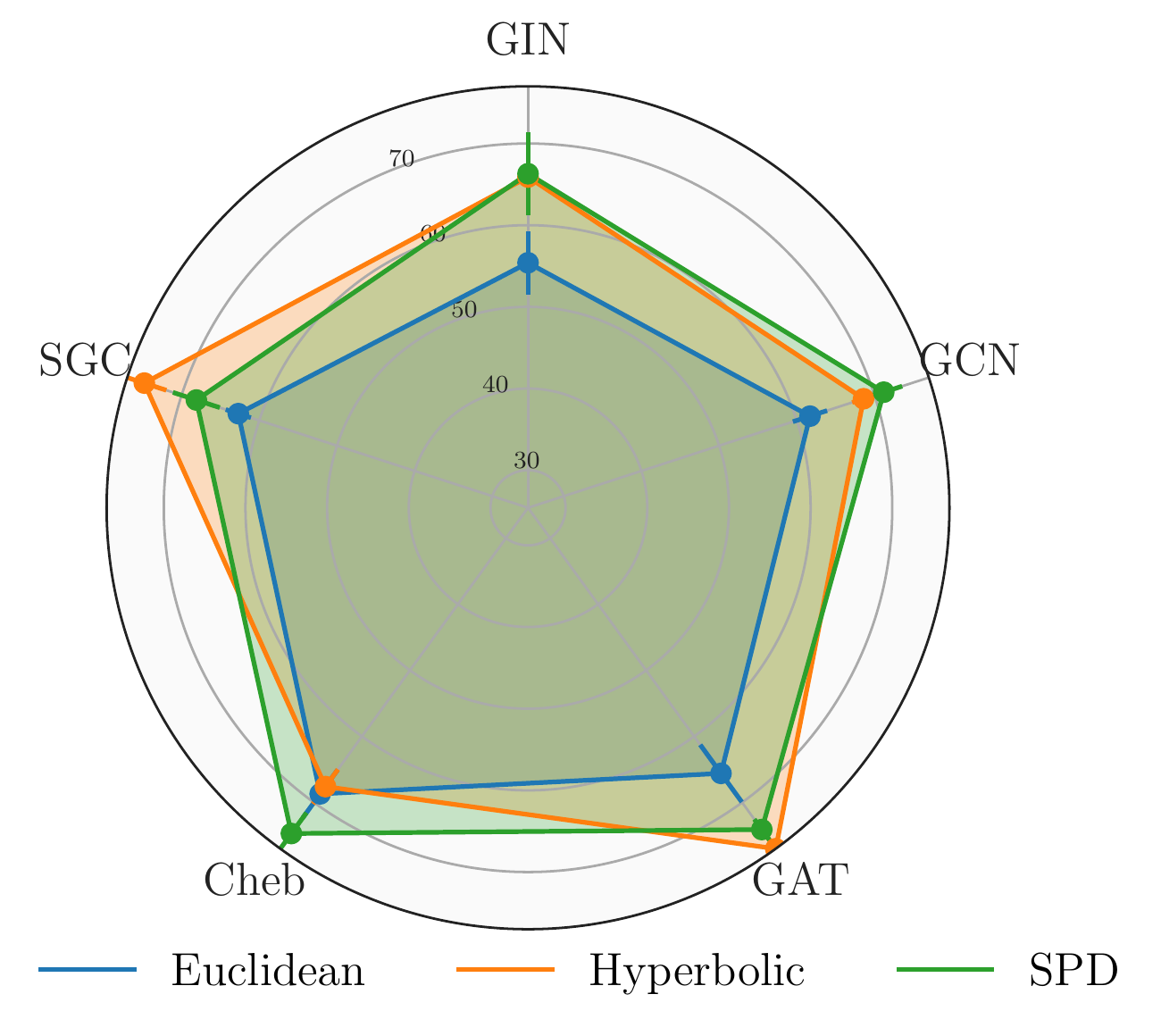}
    }
    \subfigure[Pubmed ($\delta=3.5$)]{%
    \includegraphics[width=0.3\textwidth]{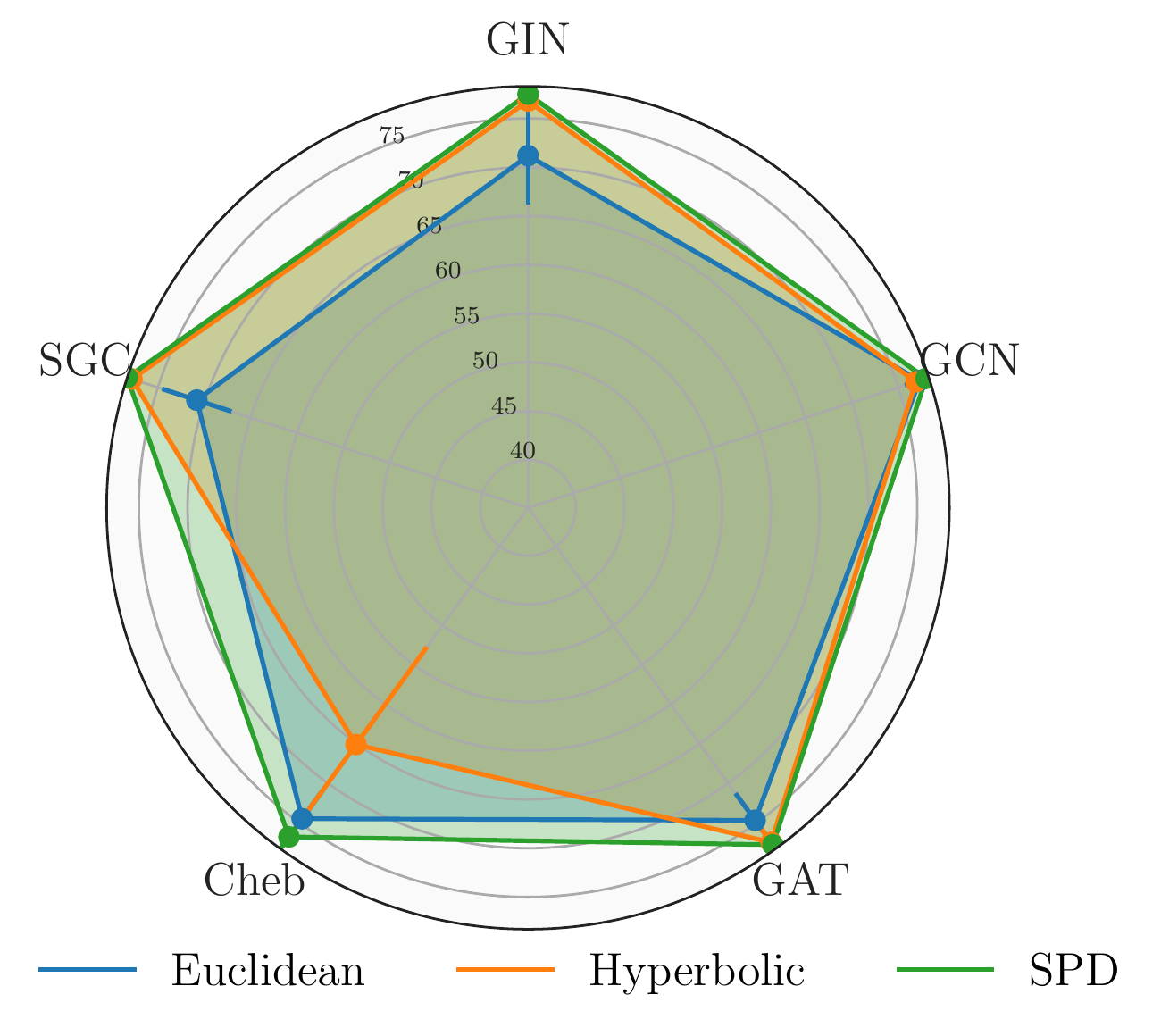}
    }    
    \subfigure[Citeseer ($\delta=5.0$)]{%
    \includegraphics[width=0.3\textwidth]{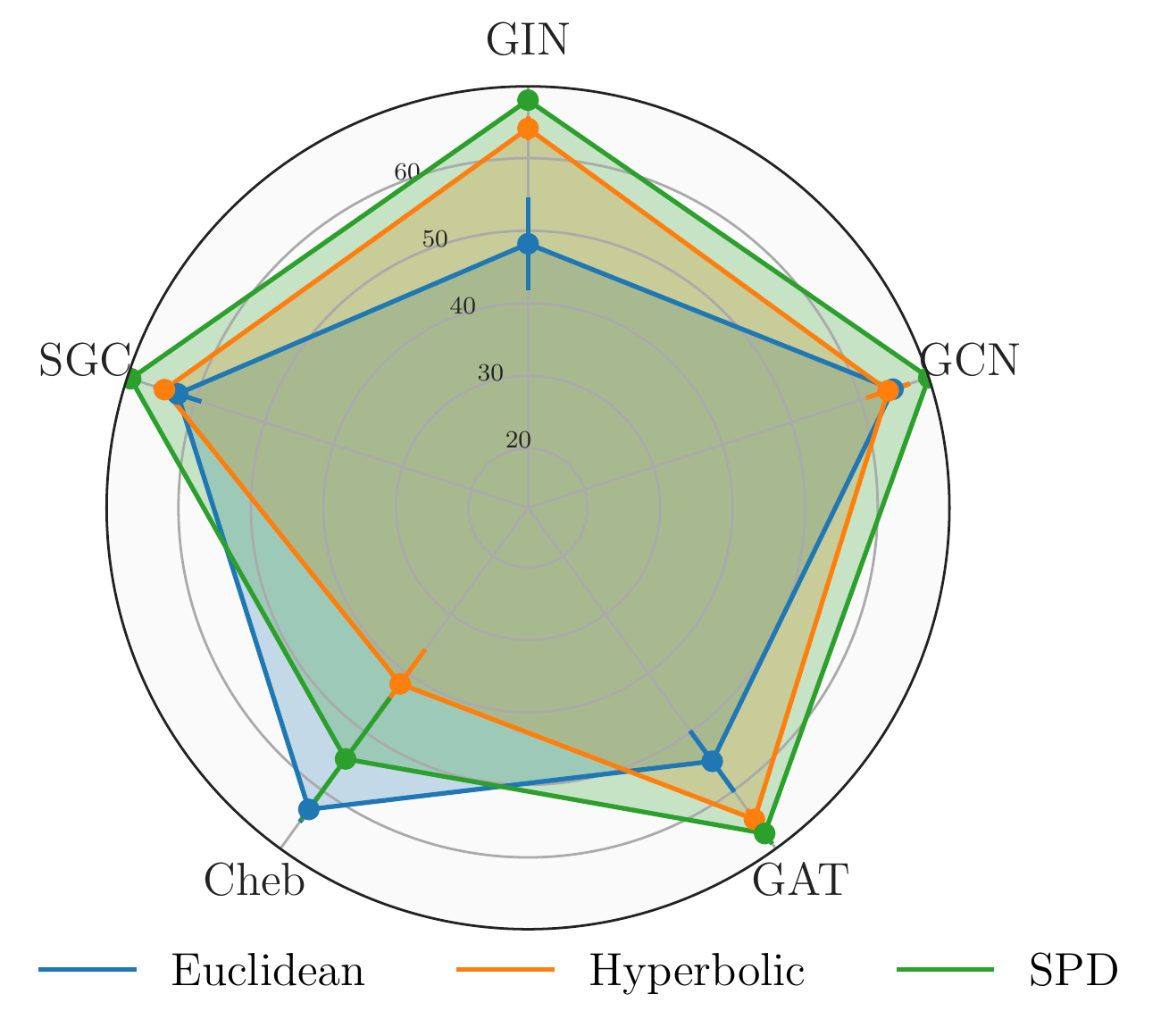}
    }    
    \subfigure[Cora ($\delta=11.0$)]{%
    \includegraphics[width=0.3\textwidth]{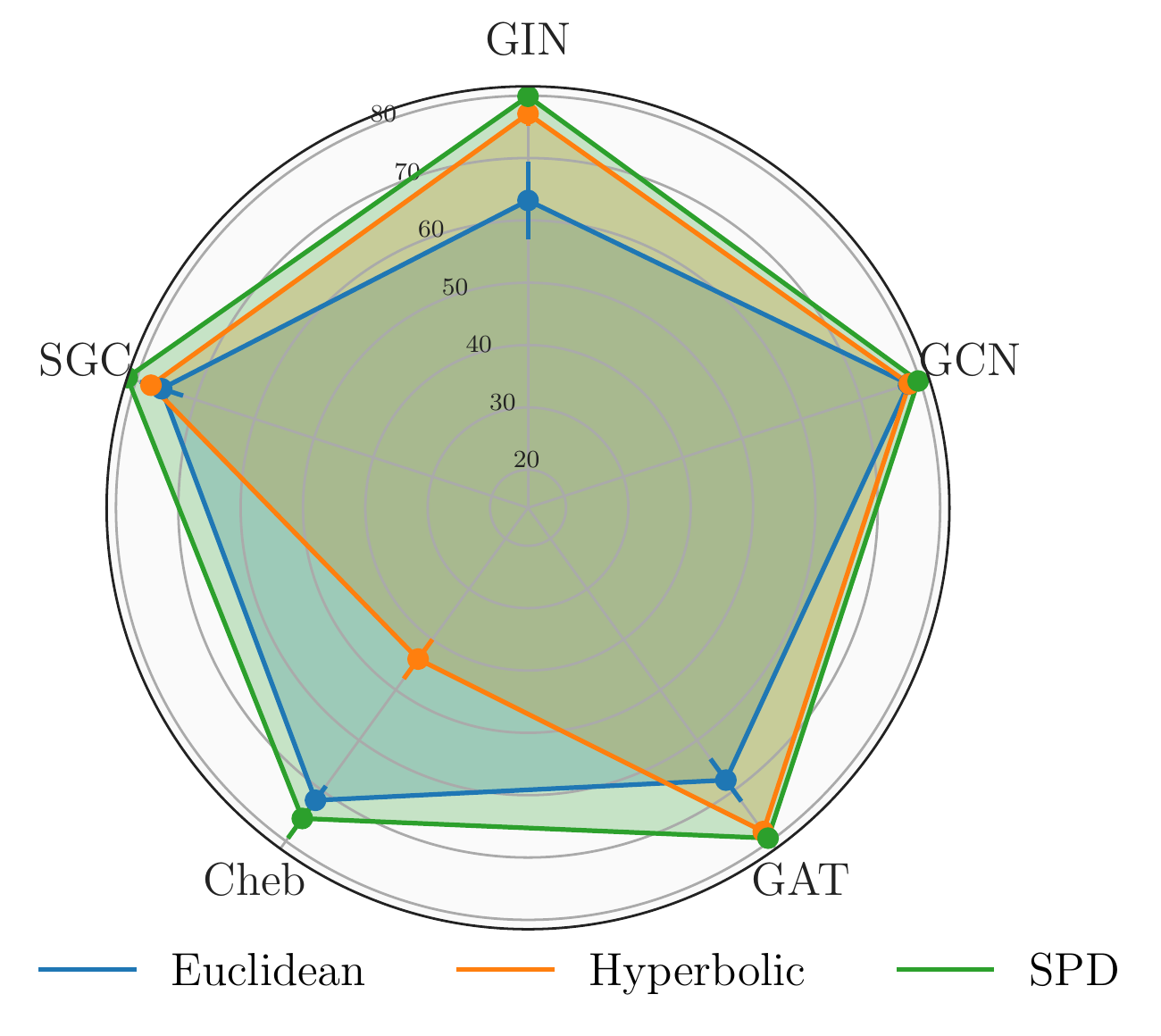}
    }        
}
{\caption{Evaluation of five graph neural networks coupled with \textsc{Linear-XE} on five node classification datasets in the three 6-dimensional spaces: $\mathbb{R}^6$, $\mathbb{H}^6$ and $\spd_3$. Each radar chart shows classification accuracy (on a varying scale, as noted by the gridlines with circular shapes) from the five GNN architectures on a dataset. Each dataset has only one graph. $\delta$-hyperbolicity shows the degree to which the dataset graph is a hyperbolic tree. A smaller $\delta$ indicates a more tree-like dataset. 
}
\label{fig:datasetsA}
}
   
\end{figure*} 
}

\newcommand{\insertRunningtime}{
\begin{figure}%
\centering
\includegraphics[width=0.6\linewidth]{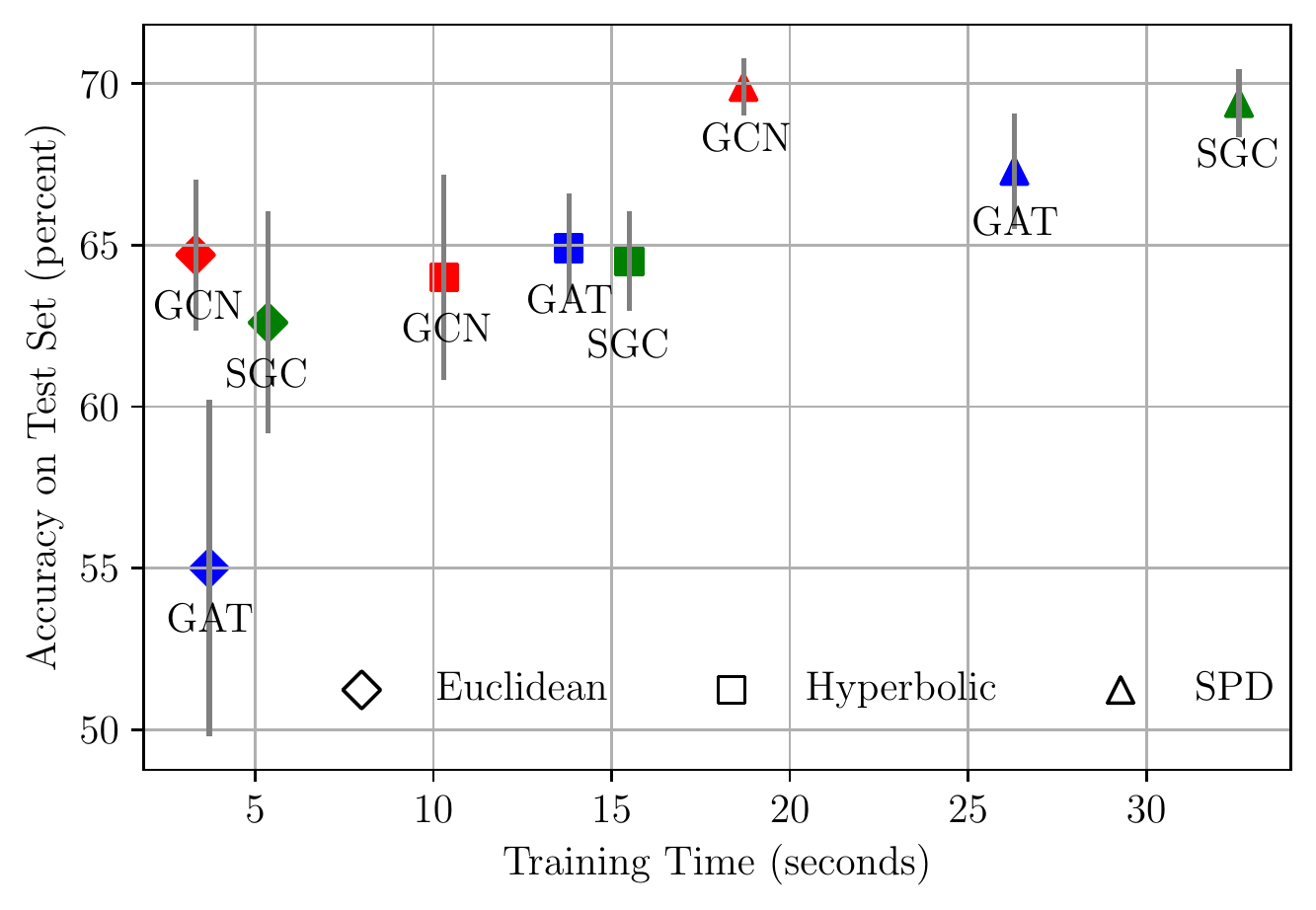}
\caption{Evaluation of three 6-dimensional spaces across graph neural networks in terms of training time and accuracy on Citeseer ($\delta = 5.0$). Each point has a unique pattern that combines color and shape. For instance, a red triangular means GCN in $\spd$.}
\label{fig:runningtime}
\end{figure}
}

\newcommand{\insertHyperbolicity}{
\begin{figure}%
\centering
\includegraphics[width=0.9\linewidth]{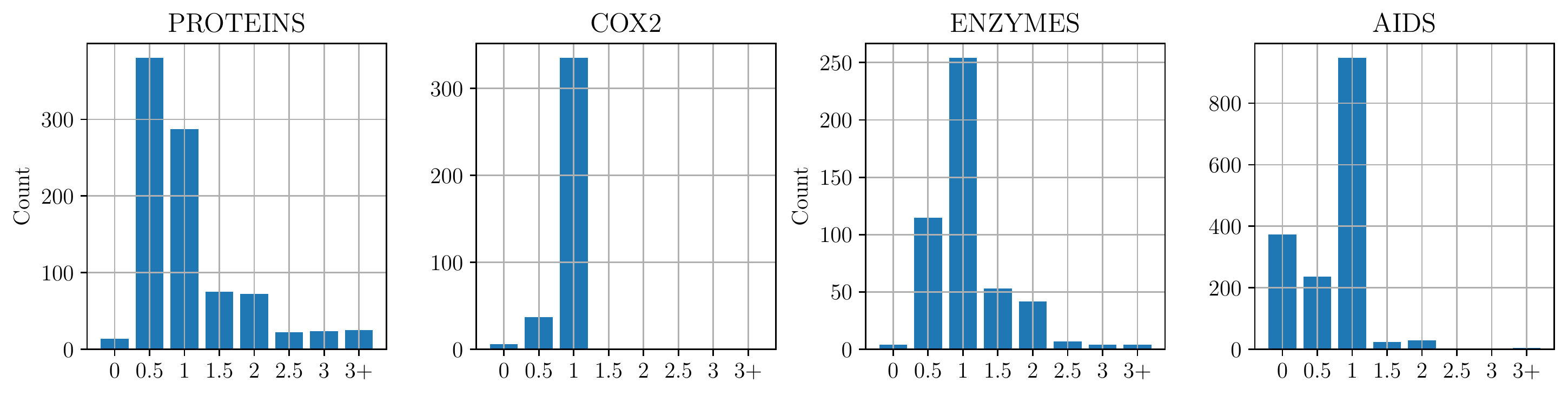}
\caption{Distributions of $\delta$-hyperbolicity on four graph classification datasets, where each instance consists of one graph. Y-axis shows the number of graphs for 
a given $\delta$-hyperbolicity on X-axis.
}
\label{fig:hyperbolicity}
\end{figure}
}

\newcommand{\insertDatasetsB}{
\begin{figure}[htbp]
\centering
\scalebox{.76}
{%
    \subfigure[COX2]{%
    \includegraphics[width=0.32\linewidth]{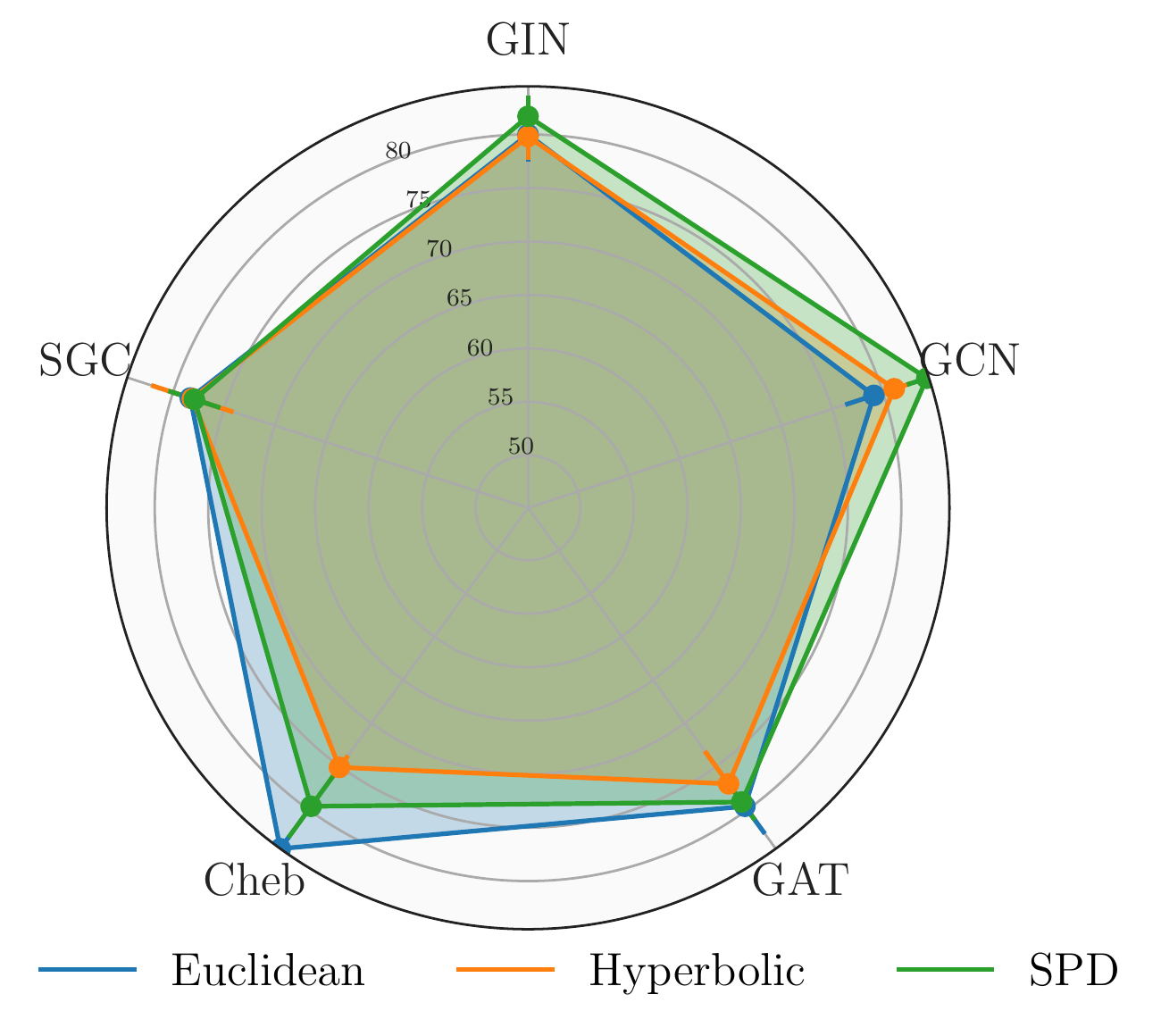}
    } 
    \subfigure[AIDS]{%
    \includegraphics[width=0.32\textwidth]{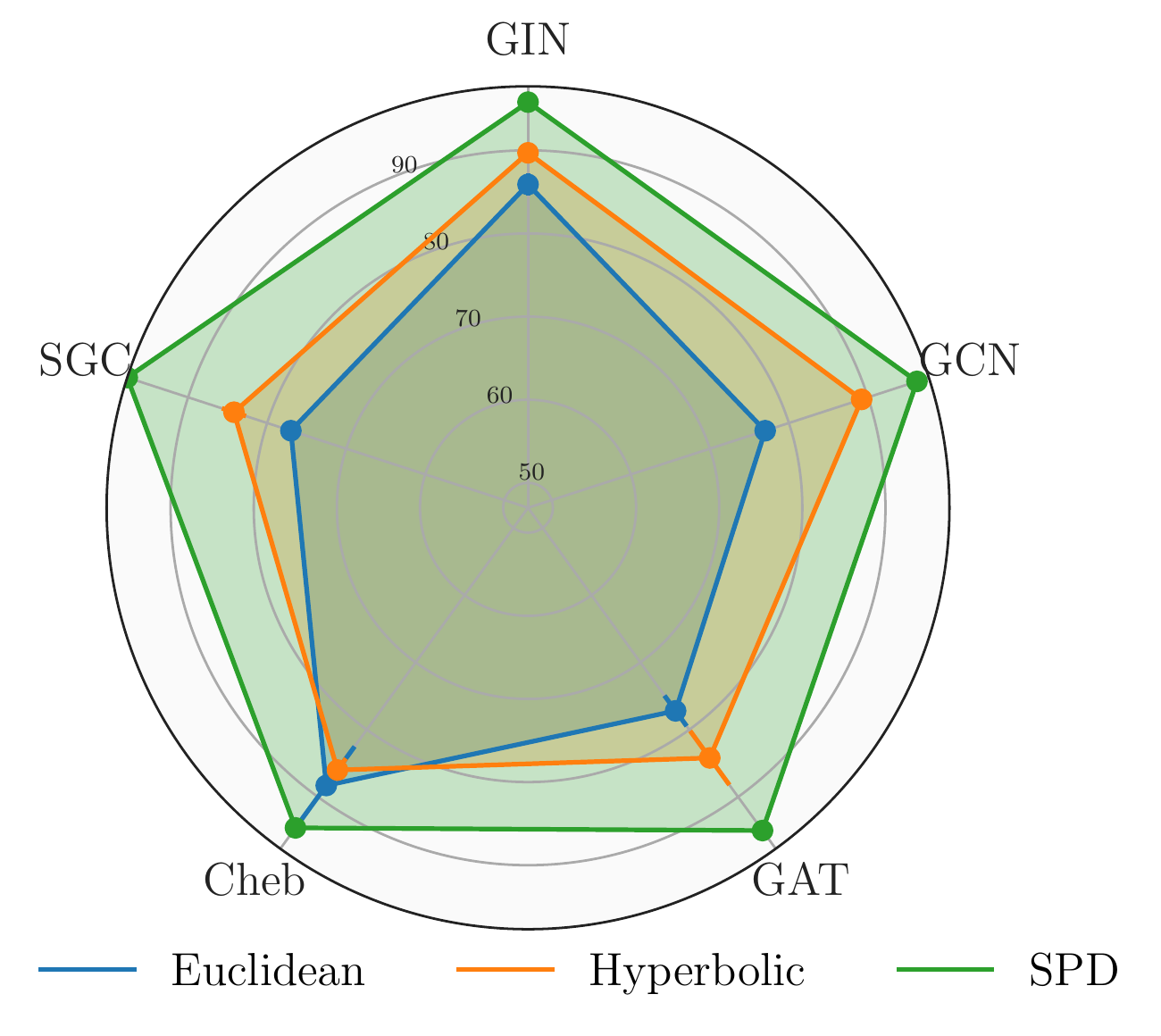}
    }
    \subfigure[ENZYMES]{%
    \includegraphics[width=0.32\textwidth]{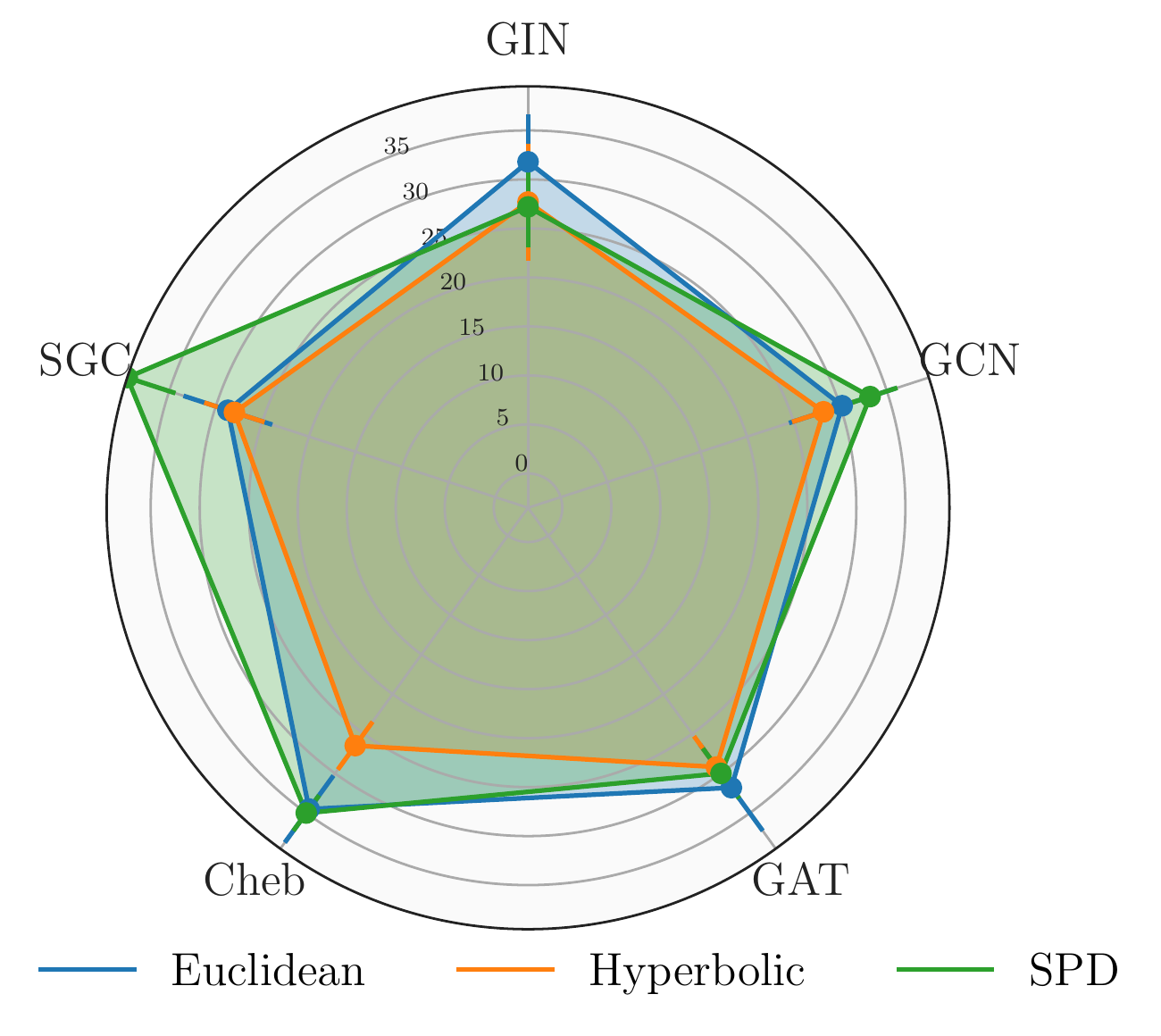}
    }    
    \subfigure[PROTEINS]{%
    \includegraphics[width=0.32\textwidth]{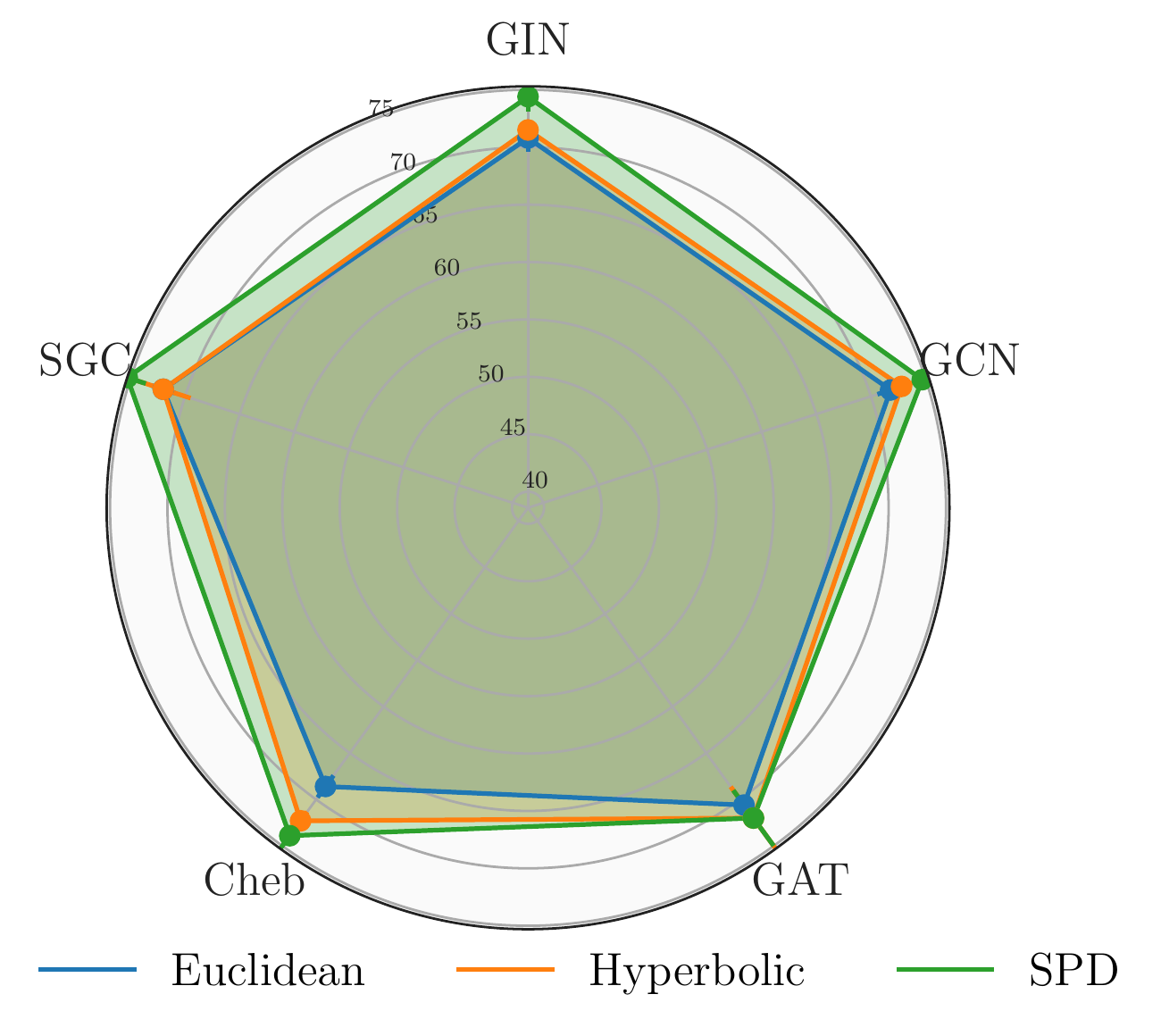}
    }    
}
{\caption{Evaluation of five graph neural networks with \textsc{Linear-XE} on four graph classification datasets in the three 36-dimensional spaces: $\mathbb{R}^{36}$, $\mathbb{H}^{36}$ and $\spd_8$. }
\label{fig:datasetsB}
}
   
\end{figure} 
}

\newcommand{\insertClassSeparation}{
\begin{figure*}[htbp]
\centering
{%
    \subfigure[Euclidean space]{%
    \includegraphics[width=0.315\linewidth]{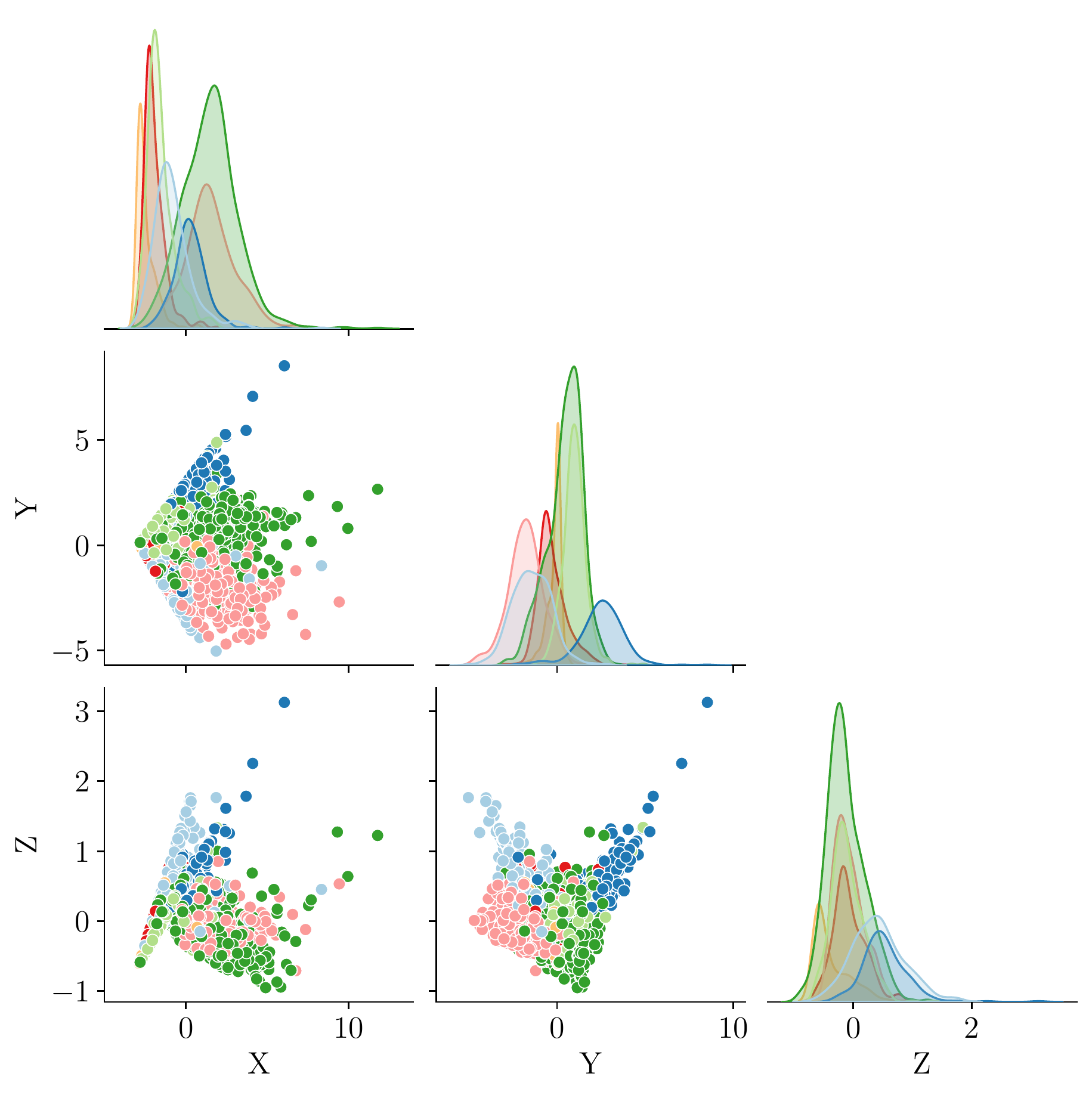}
    } 
    \subfigure[Hyperbolic space]{%
    \includegraphics[width=0.32\textwidth]{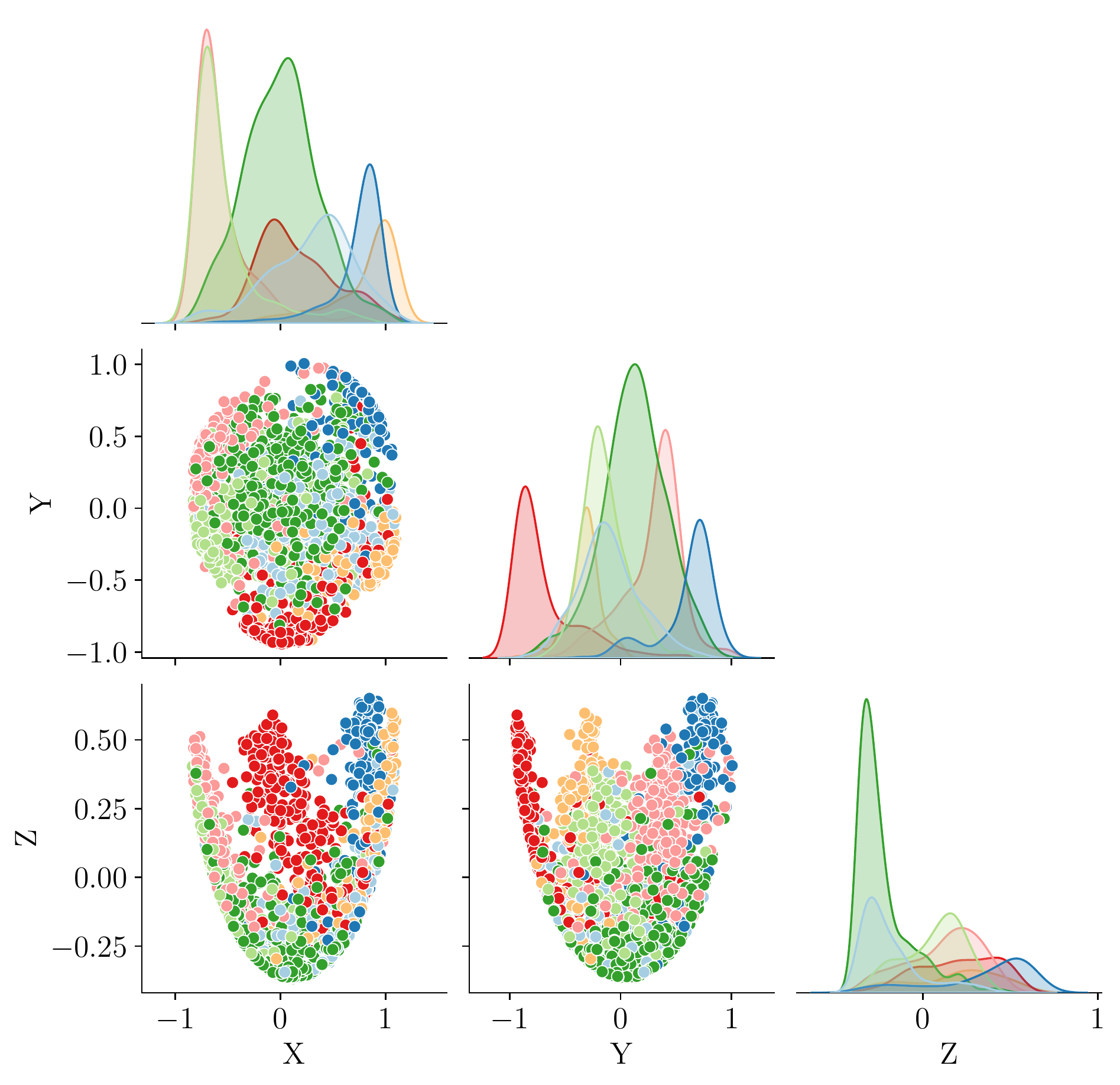}
    }
    \subfigure[SPD space]{%
    \includegraphics[width=0.32\textwidth]{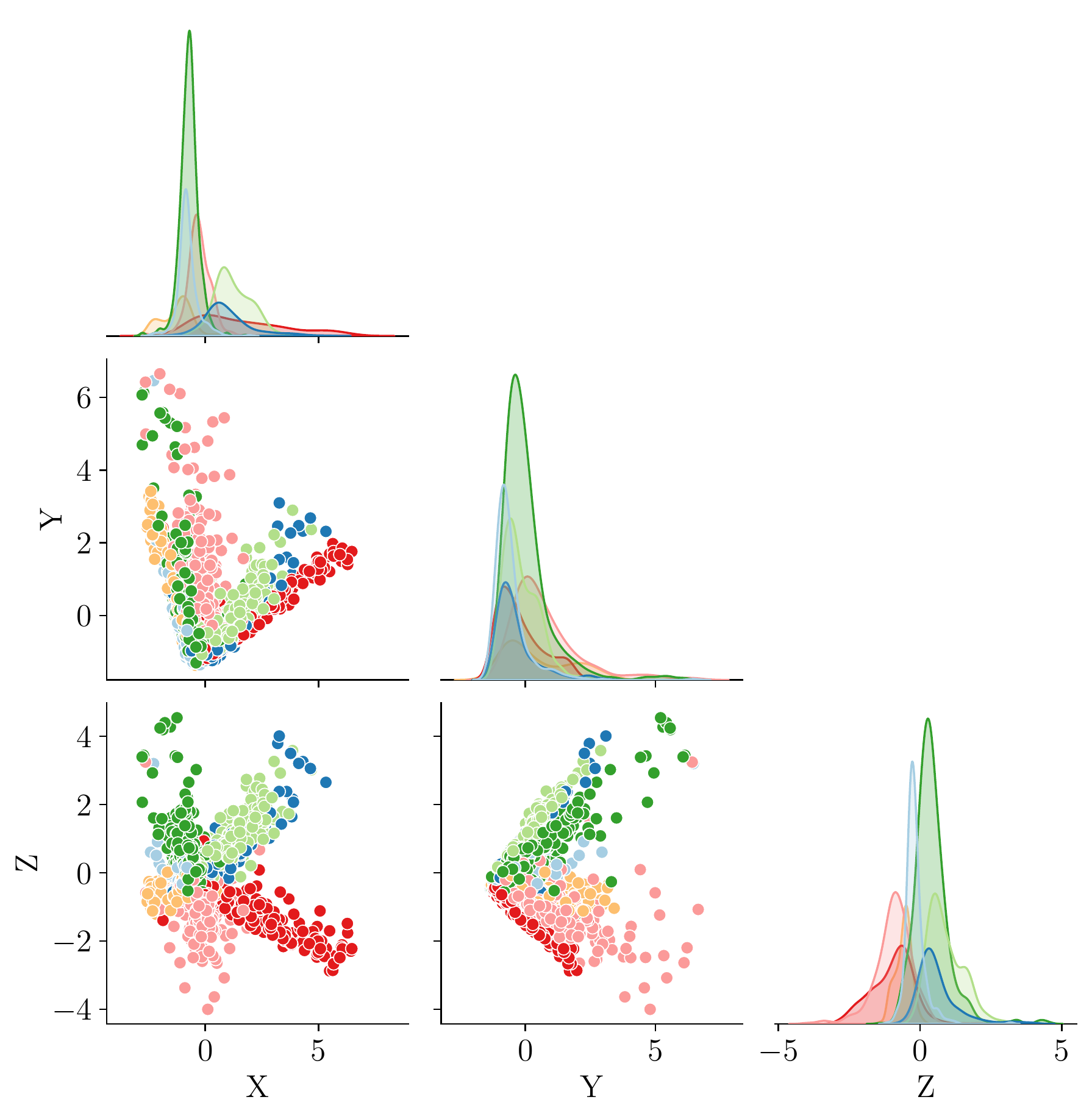}
    }    
}
{\caption{
Vizualizations of the node embeddings into three $6$-dimensional geometries for SGC on the Cora dataset. In each space, the nodes are vectorized and then projected linearly to $\RR^3$ via PCA. 
}
\label{fig:class_separation}
}
   
\end{figure*} 
}
\newcommand{\insertArchitectures}{
\begin{table*}[t] 
\centering
\scriptsize
\begin{tabular}{@{}l|c | l| l@{}}
\midrule
Operations & GNNs & Euclidean Space & Hyperbolic and SPD Space  \\ \midrule
Feature Trans & All & $h_{i}^{l} = W^l x_i^{l-1}$ & $Q_{i}^{l} = M^l \circledcirc Z_{i}^{l-1}$ \\ \midrule
\multirow{3}{*}{Propagation} & GCN & $p_{i}^{l} = \sum_{j \in \mathcal{N}(i)} k_{i, j} h_{j}^l$ & $P_{i}^{l} = \exp(\sum_{j \in \mathcal{N}(i)} k_{i, j} \log(Q_{j}^l))$  \\ 
& GAT & $p_{i}^{l} = \alpha_{i,i} h_i^l + \sum_{j \in \mathcal{N}(i)} \alpha_{i, j} h_j^l$ & $P_{i}^{l} = \exp(\alpha_{i,i}\log(Q_i) + \sum_{j \in \mathcal{N}(i)} \alpha_{i, j} \log(Q_{j}^l))$ \\ 
& Cheb & $p_{i}^{l} = h_i^l + W^l \sum_{j \in \mathcal{N}(i)} k_{i, j} h_{j}^l$ & $P_{i}^{l} = Q_{i}^{l} \oplus (M^l \circledcirc \exp(\sum_{j \in \mathcal{N}(i)} k_{i, j} \log Z_{j}^{l-1}))$ \\ \midrule
Bias\&Nonlin & All & $x_{i}^{l} = \varphi (p_{i}^l + b^l)$ & $Z_{i}^{l} = \varphi (P_{i}^{l} \oplus B^{l})$ \\ \midrule                                    
\end{tabular}
  \caption{Comparison of operations in different spaces across three graph neural networks, i.e., GCN \cite{kipf2017gnn} , GAT \cite{veličković2018graph} and 1-order Cheb \cite{defferrard2016convolutional}. SGC \cite{wu2019simplifying} and GIN \cite{xu2019gin} are presented in Appendix \ref{sec:graph_architecture}, which indeed apply propagation before feature transformation.}
  \label{tab:architectures}
\end{table*}
}

\newcommand{\insertSGCGIN}{
\begin{table*}[t] 
\caption{Comparison of operations over two graph neural networks (SGC and GIN) in different spaces. }
\centering
\footnotesize
\begin{tabular}{@{}l|c | c@{}}
\midrule
Operations & GNNs & Euclidean Space \\ \midrule
\multirow{2}{*}{Propagation} & SGC & $h_{i}^{l} = \sum_{j \in \mathcal{N}(i)} k_{i, j} \sum_{z \in \mathcal{N}(j)} k_{j, z} x_{z}^{l-1}$  \\ 
& GIN & $h_{i}^{l} = (1+\epsilon) x_i^{l-1} + \sum_{j \in \mathcal{N}(i)} k_{i,j} x_{j}^{l-1}$   \\ \midrule
Feature Transformation & All & $p_{i}^{l} = W^l h_{i}^l$   \\ \midrule
Bias\&Nonlinearity & All & $x_{i}^{l} = \varphi (p_{i}^l + b^l)$ \\ \midrule    
\end{tabular}
\begin{tabular}{@{}l|c | c@{}}
\midrule
Operations & GNNs & Hyperbolic and SPD Space \\ \midrule
\multirow{2}{*}{Propagation} & SGC & $Q_{i}^{l} = \exp(\sum_{j \in \mathcal{N}(i)} k_{i, j} \sum_{z \in \mathcal{N}(j)} k_{j, z} \log Z_{z}^{l-1})$ \\ 
& GIN & $Q_{i}^{l} = \exp((1+\epsilon)\log Z_i^{l-1} + \sum_{j \in \mathcal{N}(i)} k_{i, j} \log Z_{j}^{l-1})$  \\ \midrule
Feature Transformation & All & $P_{i}^{l} = M^l \circledcirc Q_{i}^{l-1}$  \\ \midrule
Bias\&Nonlinearity & All & $Z_{i}^{l} = \phi(P_{i}^{l} \oplus B^{l})$ \\ \midrule                                                           
\end{tabular}
  \label{tab:architectures2}
\end{table*}
}

\newcommand{\insertTableDimensionsDiseaseCora}{
\begin{table*}[t] 
  \caption{Comparison between 6- and 15-dimensional spaces across five graph neural networks and three geometries on Disease and Cora. All networks use the \textsc{Linear-XE} classifier. We bold the best accuracy in each row.}
\scriptsize
\centering
\begin{tabular}{l|l|cc|cc|cc}
 \toprule
& & $\mathbb{R}^6$ & $\mathbb{R}^{15}$ & {$\mathbb{H}^6$} & $\mathbb{H}^{15}$ & $\spd(3, \mathbb{R})$ & $\spd(5, \mathbb{R})$ \\ \toprule
\multirow{5}{*}{Disease} & GINConv  & 86.4  $\pm$ 6.9 & 91.3  $\pm$ 2.3 & 95.5  $\pm$ 0.9 & \textbf{96.3}  $\pm$ 0.8 & 91.0 $\pm$ 3.0 & 93.2 $\pm$ 2.2 \\ 
&SGConv & 87.6  $\pm$ 2.9 & 97.2  $\pm$ 1.0 & \textbf{99.9}  $\pm$ 0.1 & \textbf{99.9}  $\pm$ 0.1 & 93.9  $\pm$ 2.2  & 98.0 $\pm$ 1.0 \\ 
&ChebConv & 52.4  $\pm$ 3.9 & 51.1  $\pm$ 2.6 & 53.5  $\pm$ 3.8 & \textbf{53.7}  $\pm$ 4.8 & 53.6  $\pm$ 3.4 & 52.9 $\pm$ 4.7  \\ 
&GATConv  & 92.2  $\pm$ 8.1 & 97.0  $\pm$ 3.9 & \textbf{100.0}  $\pm$ 0.0& \textbf{100.0}  $\pm$ 0.0 & 98.1  $\pm$ 2.4 & 99.9  $\pm$ 0.2 \\ 
&GCNConv  & 88.2  $\pm$ 3.1 & 89.5  $\pm$ 1.0 & 96.9  $\pm$ 0.6& \textbf{97.8}  $\pm$ 1.2 & 95.9  $\pm$ 2.1  & 96.0  $\pm$ 3.3  \\  \midrule

\multirow{5}{*}{Cora} & GINConv  & 63.2 $\pm$ 6.2 & 74.7 $\pm$ 1.8 & 77.1 $\pm$ 1.9 & \textbf{80.6} $\pm$ 0.8 & 79.9 $\pm$ 0.6 & 81.4 $\pm$ 0.3  \\
&SGConv & 75.7 $\pm$ 3.6 & 80.7 $\pm$ 0.9 & 77.5 $\pm$ 1.5 & 79.6 $\pm$ 1.3 & 81.5 $\pm$ 0.9 & \textbf{82.1} $\pm$ 0.5 \\
&ChebConv & 71.9 $\pm$ 2.8 & 76.7 $\pm$ 1.3 & 43.9 $\pm$ 3.9 & 50.5 $\pm$ 3.7 & 75.5 $\pm$ 3.9 & \textbf{80.3} $\pm$ 1.0 \\
&GATConv  & 67.9 $\pm$ 4.2 & 74.7 $\pm$ 1.5 & 78.1 $\pm$ 1.1 & \textbf{80.3} $\pm$ 0.5 & 79.4 $\pm$ 0.9 & 79.8 $\pm$ 1.0 \\
&GCNConv  & 78.1 $\pm$ 1.7 & 80.8 $\pm$ 1.2 & 78.2 $\pm$ 1.3 & 80.7 $\pm$ 0.8 & 79.7 $\pm$ 0.9 & \textbf{82.3} $\pm$ 0.5 \\  \bottomrule
                                                
\end{tabular}
  \label{tab:dimensions-disease-cora}
\end{table*}
}

\newcommand{\insertTableNodeClassification}{
\begin{table*}[t] 
\caption{Evaluation of different GNN architectures with different classifiers in the three 6-dimensional spaces, namely $\mathbb{R}^6$, $\mathbb{H}^6$, $\spd(3, \mathbb{R})$, in the node classification task. We bold the best accuracy in each row.}
\scriptsize
\centering
\begin{tabular}{l| l| c|c|ccc}
\toprule
& & $\mathbb{R}^6$ & $\mathbb{H}^6$ & \multicolumn{3}{c}{$\spd(3, \mathbb{R})$} \\
& & \textsc{Linear-XE} & \textsc{Linear-XE} & \textsc{Linear-XE} & \textsc{SVM-MM} & \textsc{NC-MM} \\ \toprule
\multirow{5}{*}{Disease} & GINConv  & 86.4 $\pm$ 6.9    & \textbf{95.5} $\pm$ 0.9 & 91.0 $\pm$ 3.0 & 95.0 $\pm$ 0.9 & 89.4 $\pm$ 4.6 \\
 & SGConv   & 87.6 $\pm$ 2.9    & \textbf{99.9} $\pm$ 0.1 & 93.9 $\pm$ 2.2   & 95.8 $\pm$ 1.0 & 93.9 $\pm$ 1.3 \\
 & ChebConv & 52.4 $\pm$ 3.9    & 53.5 $\pm$ 3.8 & 53.6 $\pm$ 3.4   & 58.9 $\pm$ 3.7 & \textbf{61.4} $\pm$ 5.5 \\
 & GATConv  & 92.2 $\pm$ 8.1    & \textbf{100.0} $\pm$ 0.0  & 98.1 $\pm$ 2.4   & 99.6 $\pm$ 0.8 & 93.7 $\pm$ 2.6 \\
 & GCNConv  & 88.2 $\pm$ 3.1    & \textbf{96.9} $\pm$ 0.6 & 95.9 $\pm$ 2.1   & 96.0 $\pm$ 1.9 & 96.8 $\pm$ 1.7 \\ \midrule
\multirow{5}{*}{Airport} & GINConv  & 55.4 $\pm$ 3.8    & 65.9 $\pm$ 4.5 & 66.3 $\pm$ 5.1   & 68.6 $\pm$ 3.0 & \textbf{74.1} $\pm$ 2.6 \\
 & SGConv   & 62.7 $\pm$ 1.6    & \textbf{74.8} $\pm$ 2.8 & 68.1 $\pm$ 3.0   & 64.9 $\pm$ 3.5 & 71.0 $\pm$ 3.1 \\
 & ChebConv & 68.7 $\pm$ 3.6    & 67.6 $\pm$ 2.6 & 74.7 $\pm$ 3.1   & 71.7 $\pm$ 1.6 & \textbf{75.3} $\pm$ 3.2 \\
 & GATConv  & 65.6 $\pm$ 4.3    & \textbf{77.0} $\pm$ 3.3 & 74.1 $\pm$ 1.5   & 69.8 $\pm$ 2.2 & 74.4 $\pm$ 2.1 \\
 & GCNConv  & 61.7 $\pm$ 2.2    & 68.6 $\pm$ 1.4 & \textbf{71.2} $\pm$ 2.3   & 63.4 $\pm$ 3.4 & 71.0 $\pm$ 1.8 \\ \midrule
\multirow{5}{*}{Pubmed} & GINConv  & 71.2 $\pm$ 4.9    & 76.8 $\pm$ 1.0 & \textbf{77.5} $\pm$ 0.6   & 77.0 $\pm$ 1.1 & 76.9 $\pm$ 0.8 \\
 & SGConv   & 70.8 $\pm$ 3.7    & 77.8 $\pm$ 0.8 & \textbf{78.3} $\pm$ 0.5   & 78.1 $\pm$ 0.6 & 77.9 $\pm$ 0.6 \\
 & ChebConv & 74.5 $\pm$ 3.6    & 65.1 $\pm$ 12.3  & 76.8 $\pm$ 3.0 & \textbf{78.0} $\pm$ 0.8 & 77.6 $\pm$ 0.9 \\
 & GATConv  & 74.7 $\pm$ 3.4    & 77.5 $\pm$ 1.6 & 77.8 $\pm$ 0.6   & 77.2 $\pm$ 2.1 & \textbf{78.1} $\pm$ 0.5 \\
 & GCNConv  & 77.3 $\pm$ 1.5    & 76.9 $\pm$ 0.5 & 78.0 $\pm$ 0.6   & 78.2 $\pm$ 0.7 & \textbf{78.6} $\pm$ 0.7 \\ \midrule
\multirow{5}{*}{Citeseer} & GINConv  & 48.2 $\pm$ 6.3    & 64.1 $\pm$ 1.6 & \textbf{68.0} $\pm$ 1.3  & 67.3 $\pm$ 1.2 & 67.0 $\pm$ 0.8 \\
 & SGConv   & 62.6 $\pm$ 3.4    & 64.5 $\pm$ 1.5 & 69.4 $\pm$ 1.0   & \textbf{69.7} $\pm$ 0.8 & 67.9 $\pm$ 1.5 \\
 & ChebConv & 63.2 $\pm$ 2.1    & 41.8 $\pm$ 5.8 & 54.6 $\pm$ 10.4  & 61.4 $\pm$ 4.4 & \textbf{64.0} $\pm$ 2.3 \\
 & GATConv  & 55.0 $\pm$ 5.2    & 64.9 $\pm$ 1.7 & 67.3 $\pm$ 1.7   & \textbf{69.2} $\pm$ 0.7 & 68.1 $\pm$ 1.1 \\
 & GCNConv  & 64.7 $\pm$ 2.3    & 64.0 $\pm$ 3.1 & \textbf{69.9} $\pm$ 0.8   & 69.2 $\pm$ 0.8 & 68.2 $\pm$ 1.0 \\ \midrule
\multirow{5}{*}{Cora} & GINConv & 63.2 $\pm$ 6.2 & 77.1 $\pm$ 1.0   & \textbf{79.9} $\pm$ 0.6   & 79.5 $\pm$ 0.6 & 78.8 $\pm$ 1.0 \\
 & SGConv   & 75.7 $\pm$ 3.6    & 77.5 $\pm$ 1.5 & 81.5 $\pm$ 0.9   & \textbf{81.8} $\pm$ 0.3 & 81.1 $\pm$ 0.6 \\
 & ChebConv & 71.9 $\pm$ 2.8    & 43.9 $\pm$ 3.9 & 75.5 $\pm$ 3.9   & 77.9 $\pm$ 2.4 & \textbf{79.2} $\pm$ 1.3 \\
 & GATConv  & 67.9 $\pm$ 4.2    & 78.1 $\pm$ 1.1 & 79.4 $\pm$ 0.9   & \textbf{81.2} $\pm$ 1.4 & \textbf{81.2} $\pm$ 1.1 \\
 & GCNConv  & 78.1 $\pm$ 1.7    & 78.2 $\pm$ 1.3 & 79.7 $\pm$ 0.9   & \textbf{80.7} $\pm$ 0.5 & 80.2 $\pm$ 1.3 \\ \bottomrule

\end{tabular}
  
  \label{tab:results-node-classification}
\end{table*}
}

\newcommand{\insertTableGraphClassification}{
\begin{table*}[t] 
  \caption{Evaluation of different GNN architectures with different classifiers in the three 36-dimensional spaces, namely $\mathbb{R}^{36}$, $\mathbb{H}^{36}$, $\spd(8, \mathbb{R})$, in the graph classification task. We bold the best accuracy in each row.}
\scriptsize
\centering
\begin{tabular}{l| l| c|c|ccc}
\toprule
& & $\mathbb{R}^{36}$ & $\mathbb{H}^{36}$ & \multicolumn{3}{c}{$\spd(8, \mathbb{R})$} \\
& & \textsc{Linear-XE} & \textsc{Linear-XE} & \textsc{Linear-XE} & \textsc{SVM-MM} & \textsc{NC-MM} \\ \toprule
\multirow{5}{*}{COX2} & GINConv & 80.0 $\pm$ 2.5 & 79.8 $\pm$ 2.1 & 81.7 $\pm$ 1.9 & 78.7 $\pm$ 2.5 & \textbf{82.3} $\pm$ 3.0 \\
& SGConv & 78.3 $\pm$ 1.8 & 78.1 $\pm$ 4.0 & 77.9 $\pm$ 2.5 & 75.1 $\pm$ 1.9 & \textbf{79.8} $\pm$ 3.5 \\
& ChebConv & 84.5 $\pm$ 1.3 & 75.1 $\pm$ 1.3 & 79.6 $\pm$ 3.5 & 78.7 $\pm$ 3.5 & \textbf{85.3} $\pm$ 2.9 \\
& GATConv & 79.6 $\pm$ 3.1 & 77.0 $\pm$ 3.7 & 79.1 $\pm$ 2.0 & 78.7 $\pm$ 0.0 & \textbf{81.7} $\pm$ 2.5 \\
& GCNConv & 79.1 $\pm$ 2.8 & 81.1 $\pm$ 1.4 & \textbf{84.3} $\pm$ 1.9 & 74.9 $\pm$ 1.2 & 77.7 $\pm$ 4.0 \\ \midrule
\multirow{5}{*}{AIDS} & GINConv & 85.9 $\pm$ 1.3 & 89.7 $\pm$ 0.6 & \textbf{95.8} $\pm$ 1.2 & 92.3 $\pm$ 1.4 & 92.3 $\pm$ 2.6 \\
 & SGConv & 77.1 $\pm$ 0.8 & 84.2 $\pm$ 1.5 & \textbf{97.7} $\pm$ 0.9 & 94.0 $\pm$ 1.1 & 92.5 $\pm$ 0.7 \\
 & ChebConv & 88.3 $\pm$ 5.8 & 86.0 $\pm$ 1.7 & \textbf{94.6} $\pm$ 1.0 & 93.0 $\pm$ 4.6 & 94.4 $\pm$ 1.6 \\
 & GATConv & 77.2 $\pm$ 2.2 & 84.2 $\pm$ 4.0 & \textbf{95.0} $\pm$ 1.0 & 93.2 $\pm$ 4.1 & 94.2 $\pm$ 0.7 \\
 & GCNConv & 77.0 $\pm$ 0.5 & 89.2 $\pm$ 1.2 & \textbf{96.2} $\pm$ 0.4 & 93.1 $\pm$ 1.8 & 91.8 $\pm$ 0.8 \\ \midrule
\multirow{5}{*}{ENZYMES} & GINConv & \textbf{31.8} $\pm$ 4.8 & 27.7 $\pm$ 5.9 & 27.2 $\pm$ 4.1 & 29.3 $\pm$ 6.9 & 31.2 $\pm$ 6.1 \\
 & SGConv & 28.7 $\pm$ 4.7 & 28.0 $\pm$ 3.2 & 39.5 $\pm$ 5.1 & 37.2 $\pm$ 4.4 & \textbf{41.5} $\pm$ 2.6 \\
 & ChebConv & 34.5 $\pm$ 4.2 & 26.5 $\pm$ 3.0 & 35.0 $\pm$ 2.2 & 38.8 $\pm$ 5.3 & \textbf{40.0} $\pm$ 7.4 \\
 & GATConv & \textbf{31.8} $\pm$ 5.4 & 29.2 $\pm$ 3.8 & 30.0 $\pm$ 3.1 & 29.3 $\pm$ 3.2 & 29.5 $\pm$ 5.3 \\
 & GCNConv & 30.2 $\pm$ 5.7 & 28.2 $\pm$ 3.3 & 33.2 $\pm$ 2.9 & 29.0 $\pm$ 1.1 & \textbf{39.2} $\pm$ 3.2 \\ \midrule
\multirow{5}{*}{PROTEINS} & GINConv & 70.8 $\pm$ 1.2 & 71.5 $\pm$ 0.8 & \textbf{74.4} $\pm$ 1.2 & 72.3 $\pm$ 2.4 & 73.0 $\pm$ 1.6 \\
 & SGConv & 72.0 $\pm$ 1.0 & 72.0 $\pm$ 2.5 & 75.3 $\pm$ 1.7 & 74.6 $\pm$ 1.5 & \textbf{77.1} $\pm$ 1.3 \\
 & ChebConv & 68.6 $\pm$ 1.1 & 72.3 $\pm$ 0.9 & 73.9 $\pm$ 2.0 & \textbf{75.1} $\pm$ 1.6 & 74.9 $\pm$ 1.6 \\
 & GATConv & 70.6 $\pm$ 0.9 & 72.0 $\pm$ 3.3 & 72.0 $\pm$ 2.9 & 72.7 $\pm$ 3.2 & \textbf{72.8} $\pm$ 1.5 \\
 & GCNConv & 71.8 $\pm$ 1.2 & 72.8 $\pm$ 1.2 & 74.7 $\pm$ 0.9 & 74.3 $\pm$ 1.2 & \textbf{75.2} $\pm$ 1.1\\ \bottomrule
\end{tabular}

  \label{tab:results-graph-classification}
\end{table*}
}

\newcommand{\insertTableClassifiers}{
\begin{table}
  \caption{Comparison of different classifiers on Citeseer (top) and Cora (bottom). We bold the best accuracy in each row.}
\footnotesize
\centering
\begin{tabular}{l| c|ccc}
\toprule
 & $\mathbb{R}^6$ & \multicolumn{3}{c}{$\spd_3$} \\
 & \textsc{Lin-XE} & \textsc{Lin-XE} & \textsc{SVM-MM} & \textsc{NC-MM} \\ \toprule
  GIN  & 48.2 \scriptsize{$\pm$ 6.3}  & \textbf{68.0} \scriptsize{$\pm$ 1.3}  & 67.3 \scriptsize{$\pm$ 1.2} & 67.0 \scriptsize{$\pm$ 0.8} \\
  SGC   & 62.6 \scriptsize{$\pm$ 3.4}    & 69.4 \scriptsize{$\pm$ 1.0}   & \textbf{69.7} \scriptsize{$\pm$ 0.8} & 67.9 \scriptsize{$\pm$ 1.5} \\
  Cheb & 63.2 \scriptsize{$\pm$ 2.1}    & 54.6 \scriptsize{$\pm$ 10.4}  & 61.4 \scriptsize{$\pm$ 4.4} & \textbf{64.0} \scriptsize{$\pm$ 2.3} \\
  GAT  & 55.0 \scriptsize{$\pm$ 5.2}    & 67.3 \scriptsize{$\pm$ 1.7}   & \textbf{69.2} \scriptsize{$\pm$ 0.7} & 68.1 \scriptsize{$\pm$ 1.1} \\
  GCN  & 64.7 \scriptsize{$\pm$ 2.3}    & \textbf{69.9} \scriptsize{$\pm$ 0.8}   & 69.2 \scriptsize{$\pm$ 0.8} & 68.2 \scriptsize{$\pm$ 1.0} \\ \midrule
  GIN & 77.1 \scriptsize{$\pm$ 1.0}   & \textbf{79.9} \scriptsize{$\pm$ 0.6}   & 79.5 \scriptsize{$\pm$ 0.6} & 78.8 \scriptsize{$\pm$ 1.0} \\
  SGC   & 75.7 \scriptsize{$\pm$ 3.6}    & 81.5 \scriptsize{$\pm$ 0.9}   & \textbf{81.8} \scriptsize{$\pm$ 0.3} & 81.1 \scriptsize{$\pm$ 0.6} \\
  Cheb & 71.9 \scriptsize{$\pm$ 2.8}    & 75.5 \scriptsize{$\pm$ 3.9}   & 77.9 \scriptsize{$\pm$ 2.4} & \textbf{79.2} \scriptsize{$\pm$ 1.3} \\
  GAT  & 67.9 \scriptsize{$\pm$ 4.2}    & 79.4 \scriptsize{$\pm$ 0.9}   & \textbf{81.2} \scriptsize{$\pm$ 1.4} & \textbf{81.2} \scriptsize{$\pm$ 1.1} \\
  GCN  & 78.1 \scriptsize{$\pm$ 1.7}    & 79.7 \scriptsize{$\pm$ 0.9}   & \textbf{80.7} \scriptsize{$\pm$ 0.5} & 80.2 \scriptsize{$\pm$ 1.3} \\ \bottomrule
\end{tabular}
  \label{tab:classifiers}
\end{table}
}

\begin{document}

\title{Modeling Graphs Beyond Hyperbolic: \\Graph Neural Networks in Symmetric Positive Definite Matrices}
\titlerunning{Modeling Graphs Beyond Hyperbolic}

\author{{Wei Zhao}\inst{1,2}(\Letter) \and
Federico Lopez\inst{1} \and
*J. Maxwell Riestenberg\inst{2} \and 
Michael Strube\inst{1} \and 
*Diaaeldin Taha\inst{2} \and 
Steve Trettel\inst{3} 
}
\authorrunning{W. Zhao et al.}
%
\institute{
Heidelberg Institute for Theoretical Studies, Germany \\
\email{\{firstname.lastname\}@h-its.org}\\
\and
Heidelberg University, Germany \\
\email{\{mriestenberg,dtaha\}@mathi.uni-heidelberg.de}\\
\and
University of San Francisco, USA \\
\email{strettel@usfca.edu}
}

\maketitle


\begin{abstract}
Recent research has shown that alignment between the structure of graph data and the geometry of an embedding space is crucial for learning high-quality representations of the data.
The uniform geometry of Euclidean and hyperbolic spaces allows for representing graphs with uniform geometric and topological features, such as grids and hierarchies, with minimal distortion. However, real-world graph data is characterized by multiple types of geometric and topological features, necessitating more sophisticated geometric embedding spaces.
In this work, we utilize the Riemannian symmetric space of symmetric positive definite matrices ($\spd$) to construct graph neural networks that can robustly handle complex graphs. 
To do this, we develop an innovative library that leverages the $\spd$ gyrocalculus tools \cite{lopez2021gyroSPD} to implement the building blocks of five popular graph neural networks in $\spd$. Experimental results demonstrate that our graph neural networks in $\spd$ substantially outperform their counterparts in Euclidean and hyperbolic spaces, as well as the Cartesian product thereof, on complex graphs for node and graph classification tasks. We release the library and datasets at \url{https://github.com/andyweizhao/SPD4GNNs}.

\end{abstract}


\section{Introduction}
\def\thefootnote{}
\footnotetext{* These authors contributed equally to this work.}
\def\thefootnote{\arabic{footnote}}
Complex structures are a common feature in real-world graph data, where the graphs often contain a large number of connected subgraphs of varying topologies (including grids, trees, and combinations thereof). 
While accommodating the diversity of such graphs
is necessary for robust representation learning, neither Euclidean nor hyperbolic geometry alone has been sufficient \cite{krioukov2010hyperbolic}.

This inefficiency stems from geometric reasons.
Properties of the embedding space strongly control which graph topologies embed with low distortion, with simple geometries selecting only narrow classes of graphs. Euclidean geometry provides the foundational example, where its abundant families of equidistant lines allow for efficient representation of grid-like structures, but its polynomial volume growth is too slow to accommodate tree-like data. This is somewhat ameliorated by moving to higher dimensions (with faster polynomial volume growth), though with a serious trade-off in efficiency \cite{bronstein2017geometric}. An alternative is to move to hyperbolic geometry, where volume growth is exponential, providing plenty of room for branches to spread out for the isometric embedding of trees \cite{buyalo2005embedding,sonthalia2020tree}.
However, hyperbolic geometry has a complementary trade-off: it does not contain equidistant lines, which makes it unfit for embedding the grid-like structures Euclidean space excelled at \cite{cannon1997hyperbolic}.

\begin{figure}[!t]
    \centering
    \includegraphics[width=0.45\columnwidth,keepaspectratio]{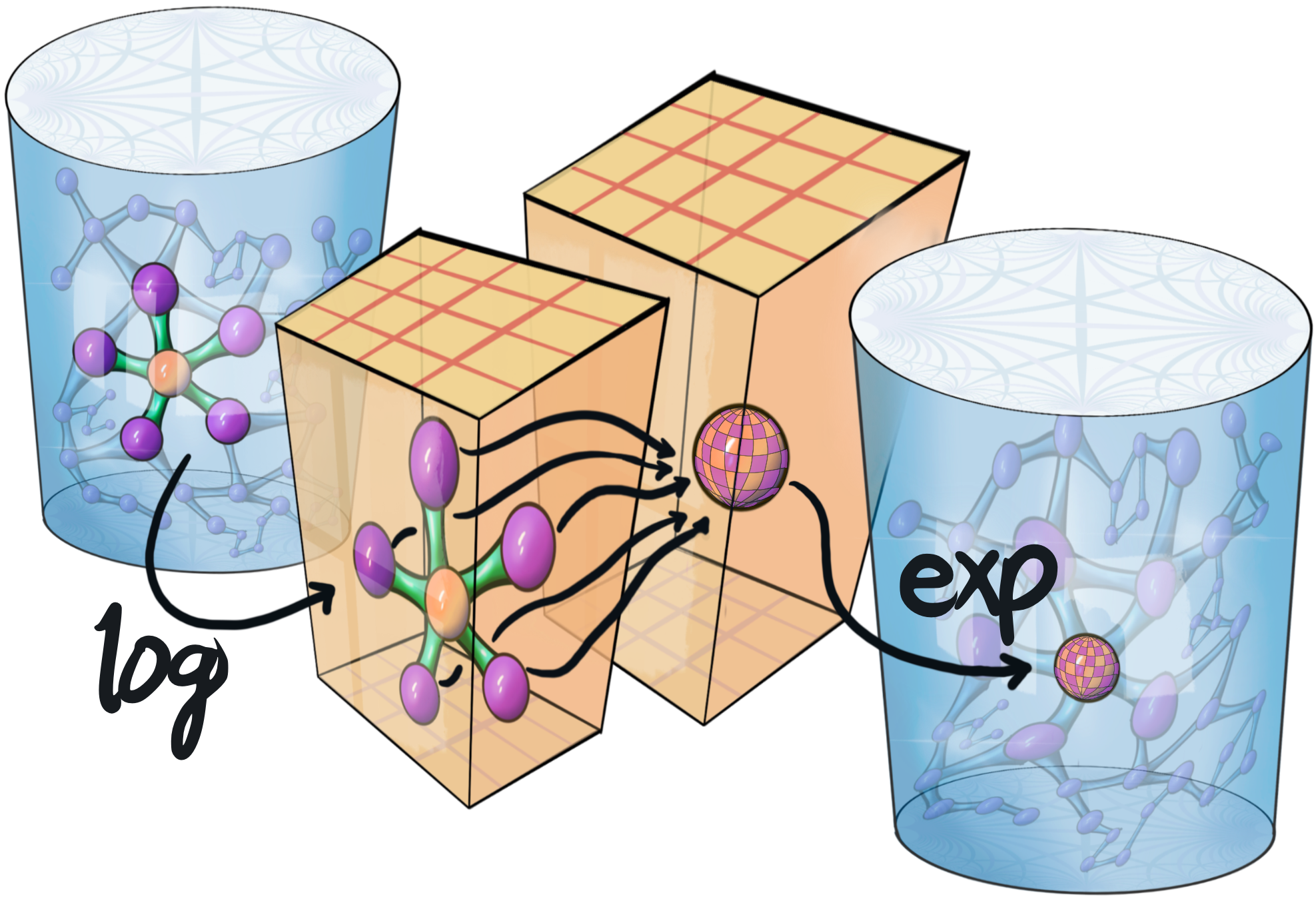}
    \caption{Propagation schema for utilizing $\spd$ geometry while performing calculations in the tangent (Euclidean) space: starting from an $\spd$ embedding, map a node and its neighbors to the tangent space via the logarithm, and perform a modified Euclidean aggregation (Table \ref{tab:architectures}) before returning to $\spd$ via the Riemannian exponential map.}
    \label{fig:label3}
    \vspace{0mm}
\end{figure}

Many proposed graph neural networks utilize representations of graph data to perform machine learning tasks in various graph domains, such as social networks, biology and molecules \cite{kipf2017gnn,veličković2018graph,chami2019hgcnn,Defferrard2020DeepSphere,satorras2021n,rusch2022graph,di2022graph,barcelo2022weisfeiler}. A subset of these networks take seriously
the constraints of geometry on representation capability, and work to match various non-Euclidean geometries to common structures seen in graph data. For instance, Chami et al. \cite{chami2019hgcnn} and Defferrard et al. \cite{Defferrard2020DeepSphere} show that constructing graph neural networks in hyperbolic and spherical spaces have been successful in embedding graphs with hierarchical and cyclical structures, respectively.  However, the relative geometric simplicity of these spaces poses serious limitations including (a) the need to know the graph structure prior to choosing the embedding space, and (b) the inability to perform effectively with graphs built of geometrically distinct sub-structures, a common feature of real-world data.

Avoiding these limitations may necessitate resorting to more complex geometric spaces. For example, Gu et al. \cite{gu2019lmixedCurvature} employed Cartesian products of various geometric spaces to represent graphs with mixed geometric structures. But any such choice must be carefully considered: isometries play an essential role in the construction of the above architectures, and any increase in complexity accompanied by too great a decrease in symmetry may render a space computationally intractable.

Riemannian symmetric spaces, which have a rich geometry encompassing all the aforementioned spaces, 
strike an effective balance between geometric generality and ample symmetry. Lopez et al. \cite{lopez2021symmetric} proposed particular symmetric spaces, namely Siegel spaces, 
for graph embedding tasks, and demonstrated that many different classes of graphs embed in these spaces with low distortion. Lopez et al. \cite{lopez2021gyroSPD} suggested utilizing the symmetric space $\spd$ of symmetric positive definite matrices that is less computationally expensive than Siegel spaces. Furthermore, they developed gyrocalculus tools that enable ``vector space operations'' on $\spd$.

Here we extend the idea of Lopez et al. \cite{lopez2021gyroSPD} to construct graph neural networks in $\spd$, particularly by utilizing their gyrocalculus tools 
to implement the building blocks of graph neural networks in $\spd$. The building blocks include (a) feature transformation via isometry maps, (b) propagation via graph convolution in the tangent space of $\spd$ (the space of symmetric matrices $S_n$), (c) bias addition via gyrocalculus, (d) non-linearity acting on eigenspace, and (e) three classification layers. We develop \textit{SPD4GNNs}, an innovative library that showcases training five popular graph neural networks in $\spd_n$, alongside the functionality for training them in Euclidean and hyperbolic spaces.

We perform experiments to compare four ambient geometries (Euclidean, hyperbolic, products thereof, and $\spd$) 
across popular graph neural networks, evaluated on the node and graph classification tasks on nine datasets with varying complexities. Results show that constructing graph neural networks in $\spd$ space leads to big improvements in accuracy over Euclidean and hyperbolic spaces on complex graphs, at the cost of doubling (resp.\ quadrupling) the training time of graph neural networks compared to hyperbolic space (resp.\ Euclidean space).

Finally, we provide a summary of the numerical issues we encountered and the solutions to addressing them (see Appendix \ref{sec:numerical_issues}).

\section{Related Work}
\label{sec:related-work}

\paragraph{Graph Neural Networks.}
Graph neural networks (GNNs) have been profiled as the de facto solutions for learning graph embeddings \cite{kipf2017gnn,veličković2018graph,wu2019simplifying,rusch2022graph,di2022graph,barcelo2022weisfeiler}. These networks can be differentiated into two dimensions: (a) how they propagate information over graph nodes and (b) which geometric space they use to embed the nodes. 
In \textbf{Euclidean space}, a class of GNNs has been proposed that represents graph nodes in a flat space and propagates information via graph convolution in various forms, such as using Chebyshev polynomial filters \cite{defferrard2016convolutional,kipf2017gnn}, high-order filters \cite{wu2019simplifying}, importance sampling \cite{sage:2017}, attention mechanisms \cite{veličković2018graph}, graph-isomorphism designs \cite{xu2019gin,satorras2021n,barcelo2022weisfeiler}, and differential equations \cite{eliasof2021pde,rusch2022graph,di2022graph}. In contrast, \textbf{non-Euclidean spaces} have a richer structure for representing geometric graph structures in curved spaces. Recently, there has been a line of GNNs developed in these spaces that performs graph convolution on different Riemannian manifolds in order to accommodate various graph structures, such as hyperbolic space on tree-like graphs \cite{chami2019hgcnn,hgnn:2019,yu2022hyla}, spherical space on spherical graphs \cite{Defferrard2020DeepSphere}, and Cartesian products of thereof \cite{gu2018learning}.

\paragraph{$\spd$ Space.} 

Representing data with SPD matrices has been researched for many years, with the representations being primarily in the form of covariance matrices \cite{dong2017spdToFaceRecog,huang2017riemannianNetForSPDMatrix,zhang2018manifoldToManifold,gao2019robustRepreSPD,chakraborty2020manifoldNet}. These matrices capture the statistical dependencies between Euclidean features. Recent research  focused on designing the building blocks of neural networks in the space of covariance matrices. This includes feature transformation that maps Euclidean features to covariance matrices via geodesic Gaussian kernels \cite{dong2017spdToFaceRecog,brooks2019riemannianRadarData}, nonlinearity on the eigenvalues of covariance matrices \cite{huang2017riemannianNetForSPDMatrix}, convolution through $\spd$ filters \cite{zhang2018manifoldToManifold} and Frech\'et mean \cite{chakraborty2020manifoldNet}, Riemannian recurrent networks \cite{chakraborty2018recurrentSPD}, and Riemannian batch normalization \cite{brooks2019riemBNforSPD}.

Nguyen et al. \cite{nguyen2019spdHandGestureRecog} recently approached hand gesture classification by embedding graphs into $\spd$ via a neural network. The architectures we consider here are different. While we alternate between $\spd$ and its tangent space using the exponential and logarithm maps, Nguyen et al. \cite{nguyen2019spdHandGestureRecog} do so via an aggregation operation and the log map. Further, we couple this alternation with our building blocks to operate graph neural networks in $\spd$.

\section{Background}
\label{sec:background}

\subsection{The Space $\spd$}
\begin{figure*}[!t]
    \centering
    \includegraphics[width=0.8\textwidth,keepaspectratio]{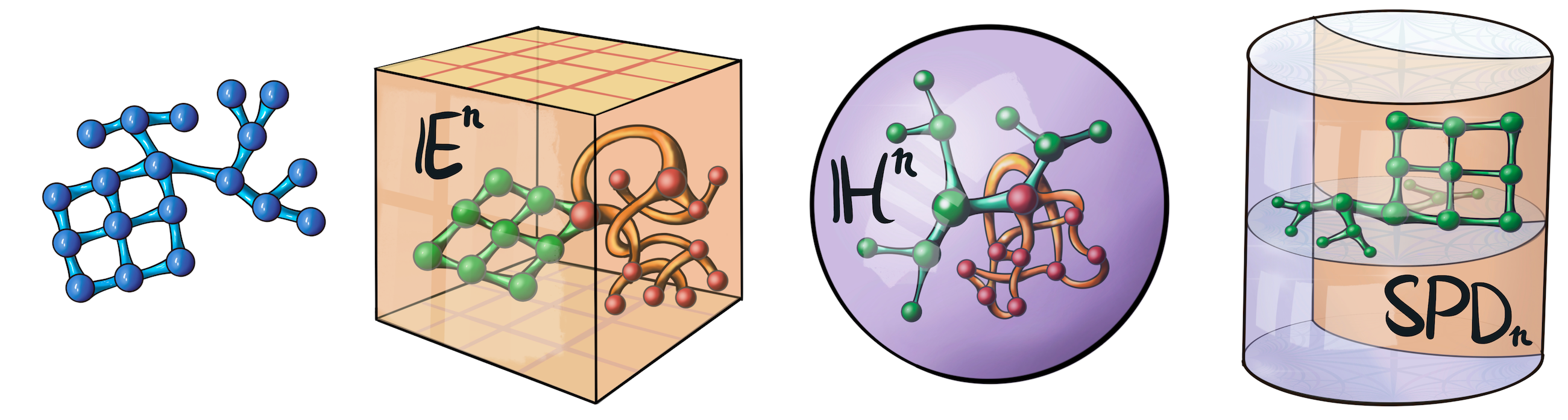}
    \caption{Graphs exhibiting both euclidean/grid-like and hyperbolic/tree-like features (left) cannot embed well in either euclidean or hyperbolic spaces due to the impossibility of isometrically embedding trees in Euclidean spaces and grids in hyperbolic spaces (center). However, $\spd_n$ (right) contains both euclidean and hyperbolic subspaces, which allows embedding a broad class of graphs, including the example in the figure.}
    \label{fig:label2}
    \vspace{0mm}
\end{figure*}

We let $\spd_n$ denote the space of positive definite real symmetric $n \times n$ matrices. This space has the structure of a Riemannian manifold of non-positive curvature of $n(n+1)/2$ dimensions. The tangent space to any point of $\spd_n$ can be identified with the vector space $S_n$ of all real symmetric $n\times n$ matrices. $\spd_n$ is more flexible than Euclidean or hyperbolic geometries, or products thereof. In particular, it contains $n$-dimensional Euclidean subspaces, $(n-1)$-dimensional hyperbolic subspaces, products of {$\lfloor\frac{n}{2}\rfloor$} hyperbolic planes, and many other interesting spaces as totally geodesic submanifolds; see the reference \cite{helgason1078diffGeom} for an in-depth introduction to these well-known facts. While it is not yet fully understood how our proposed models leverage the Euclidean and hyperbolic subspaces in $\spd_n$, we hypothesize that the presence of these subspaces is an important factor in the superior performance of $\spd_n$ graph neural networks. Refer to Figure \ref{fig:label2} for a demonstration of how this hypothesis may manifest. 

\paragraph{Riemannian Exponential and Logarithmic Maps.} 
For $\spd_n$, the Riemannian exponential map at the basepoint $I_n$ agrees with the standard matrix exponential $\exp\colon S_n\to \spd_n$. This map is a diffeomorphism with inverse the matrix logarithm $\log\colon \spd_n\to S_n$. These maps allow us to pass from $\spd_n$ to $S_n$ and back again. Given any two points $X,Y\in \spd_n$, there exists an isometry (i.e., a distance-preserving transformation) that maps $X$ to $Y$. As such, the choice of a basepoint for the exponential and logarithm maps is arbitrary since any other point can be mapped to the basepoint by an isometry. In particular, there is no loss of generality with fixing the basepoint $I_n$ as we do.

\paragraph{Non-linear Activation Functions in $\spd_n$.} We use two non-linear functions on SPD matrices, namely (i) ReEig \cite{huang2017riemannianNetForSPDMatrix}: factorizing a point $P \in \spd_n$ and then employing the ReLU-like
non-linear activation function $\varphi_a$ to suppress the positive eigenvalues that are bigger than 0.5
\footnote[2]{TgReEig equals ReEig in the case of $\varphi_a(x) = \max(x, 1)$. 
}:
\begin{equation}
    \varphi^{\spd}(P) = U \varphi_a(\Sigma) U^T\hspace{1cm}P = U \Sigma U^T
\end{equation}
(ii) TgReEig: projecting $P \in \spd_n$ into the tangent space and then suppressing the negative eigenvalues of the projected point $\in S_n$ with the ReLU non-linear activation function $\varphi_b$, i.e. $\varphi^{\spd}(P) = U \exp(\varphi_b(\log(\Sigma))) U^T$.

\subsection{Gyrocalculus on SPD}
\label{subsec:gyro-spd}

\paragraph{Addition and Subtraction.} 
Gyro-calculus is a way of expressing natural analogues of vector space operations in Riemannian manifolds. Following Lopez et al. \cite{lopez2021gyroSPD}, given two points $P,Q\in \spd_n$, we denote gyro-addition and gyro-inversion by:
\begin{equation}
\label{eq:gyro-addition}
      P \oplus Q =\sqrt{P}Q\sqrt{P}\hspace{1cm}\ominus P = P^{-1}
\end{equation}
For $P,Q\in\spd_n$, the value $P\oplus Q \in \spd_n$ is the result of applying the $\spd_n$-translation moving the basepoint $I_n$ to $P$, evaluated on $Q$. 

\paragraph{Isometry Maps.}
Any invertible $n \times n$ real matrix $M \in \mathrm{GL}(n,\RR)$ defines an isometry of $\spd_n$ by
\begin{equation}
    \label{eq:rotref}
    M \circledcirc P = MPM^T
\end{equation}
where $P \in \spd_n$. 

Lopez et al. \cite{lopez2021gyroSPD} proposed defining $M$ in two forms, namely a rotation element in $SO(n)$ and a reflection element in $O(n)$.
In this case, the choice of rotation and reflection becomes a hyperparameter for $M$, and that needs to be selected before training. 
In contrast, the form of $M$ we considered is more flexible and can be automatically adjusted by training data. To do so, we first let $M$ denote the orthogonal basis of a learnable square matrix, and then tune the square matrix from training data.
Thus, $M$, as an orthogonal matrix that extends rotations and reflections, is better suited to fit the complexity of graph data.

\section{Graph Neural Networks}
In this section, we introduce the notation and building blocks of graph neural networks using graph convolutional network (GCN) \cite{kipf2017gnn} as an example, and present modifications for operating these building blocks in $\spd$. Table \ref{tab:architectures} establishes parallels between five popular graph neural networks in Euclidean, hyperbolic and $\spd$ spaces.

\subsection{GCN in Euclidean Space}
\insertArchitectures

\label{sec:gnn}

Given a graph $\mathcal{G} = (\vertices,\edges)$ with a vertex set $\vertices$ and an edge set $\edges$, we define $d$-dimensional input node features $(x_{i}^{0})_{i \in \vertices}$, where the superscript $0$ indicates the first layer. The goal of a graph neural network is to learn a mapping denoted by:
\begin{equation*}
    f: (\vertices, \edges, (x_{i}^{0})_{i \in \vertices}) \rightarrow \mathcal{Z} \in \RR^{|\vertices| \times d}
\end{equation*}
where $\mathcal{Z}$ is the space of node embeddings obtained from the final layer of GCN, which we take as the input of classification layer to perform downstream tasks.

Let $\neighs(i) = \{j : (i, j) \in \edges \} \cup \{i\}$ be the set of neighbors of $i \in \vertices$ with self-loops, and $(W^l, b^l)$ be a matrix of weights and a vector of bias parameters at layer $l$, and $\varphi(\cdot)$ be a non-linear activation function. We now introduce \textbf{message passing}, which consists of the following three components for exchanging information between the node $i$ and its neighbors at layer $l$:
\begin{align}
    \label{eq:feat-agg} h_{i}^{l} &= W^l x_{i}^{l-1} & \text{\footnotesize{Feature transform}} \\
    \label{eq:propagation} p_{i}^{l} &= \sum_{j \in \neighs(i)} k_{i,j} h_{j}^l & \text{\footnotesize{Propagation}} \\
    \label{eq:bias-and-nonlin} x_{i}^{l} &= \varphi(p_{i}^{l} + b^{l}) & \text{\footnotesize{Bias \& Nonlinearity}}
\end{align}
where $k_{i,j} = c_i^{-\frac{1}{2}} c_j^{-\frac{1}{2}}$ with $c_i$ as the cardinality of $\neighs(i)$.
$k_{i,j}$ represents the relative importance of the node $j$ to the node $i$.

\subsection{GCN in $\spd$}

\paragraph{Mapping from Euclidean to SPD space.} Oftentimes, input node features are not given in $\spd$, but in Euclidean space $(x_{i}^{0})_{i \in \vertices} \in \RR^d$. Therefore, we design a transformation that maps Euclidean features to a point in $\spd$. 
To do so, we first learn a linear map that transforms the $d$-dimensional input features into a vector of dimension $n(n + 1)/2$, that we arrange as the upper triangle of an initially zero matrix $A \in \RR^{n \times n}$.
We then define a symmetric matrix $U \in S_n$ such that $U = A + A^T$.
We now apply the exponential map such that $Z = \exp(U)$, which moves the coordinates from the tangent space $S_n$ to the original manifold $\spd_n$. 
Thus, the resulting node embeddings $(Z_{i}^{0})_{i \in \vertices}$ are in $\spd_n$. 
By performing this mapping only once, we enable GNNs to operate in $\spd_n$.

\paragraph{Feature Transform.} 
We apply isometry maps to transform points in $\spd$ at different layers, denoted by: $Q_{i}^l = M^l \circledcirc Z_i^{l - 1}$, where $Q_{i}^l, Z_{i}^{l - 1} \in \spd_n$ and $M^l$ is a isometry map (see \S\ref{subsec:gyro-spd}) at layer $l$ of the GNN.

\paragraph{Propagation.} This step aggregates information from all the neighbors $\neighs(i)$ of a given node $i$, with the information weighted by the importance of a neighbor to the node $i$ (see Eq.~\ref{eq:propagation}).
We note that propagation involves addition and scaling operators. This results in two alternative approaches for computing propagation: (a) employing gyro-addition to aggregate information over the neighbors for each node; (b) computing the Riemannian Fr\'echet mean in $\spd_n$---which requires hundreds of iterations to find a geometric center.
Therefore, these approaches are costly to compute and also involve the use of cumbersome Riemannian optimization algorithms (see Appendix \ref{sec:training_details} for optimization). Here we perform aggregation via graph convolution in the space of symmetric matrices $S_n$ denoted by: $P_{i}^{l} = \exp(\sum_{j \in \neighs(i)} k_{i, j} \log(Q_{j}^l))$, where $P_{i}^{l} \in \spd_n$ and $k_{i, j}=c_i^{-\frac{1}{2}}c_j^{-\frac{1}{2}}$ (as in the Euclidean case). This is similar to the approach of Chami et al. \cite{chami2019hgcnn} by performing propagation in the tangent pace and the posterior projection through the exponential map.

\paragraph{Bias Addition and Non-linearity.} Finally, we add the bias $B^l$ at layer $l$ to the result of propagation through gyro-addition followed by applying a non-linear function, denoted by: $Z_{i}^{l} = \varphi^{\spd}(P_{i}^{l} \oplus B^l)$, with $Z_{i}^{l} \in \spd_n$ as the new embedding for the node $i$ at layer $l$ and $B^l \in \spd_n$.

\paragraph{Message Passing in $\spd$.}We establish a one-to-one correspondence between the Euclidean and SPD versions of GCN at layer $l$ for node $i$:
\begin{align}
    \label{eq:spd-feat-agg} Q_{i}^{l} &= M^l \circledcirc Z_{i}^{l-1} & \text{\footnotesize{Feature transform}} \\
    \label{eq:spd-propagation} P_{i}^{l} &= \exp(\sum_{j \in \neighs(i)} k_{i, j} \log(Q_{j}^l)) & \text{\footnotesize{Propagation}} \\
    \label{eq:spd-bias-and-nonlin} Z_{i}^{l} &= \varphi^{\spd}(P_{i}^{l} \oplus B^{l}) & \text{\footnotesize{Bias \& non-lin}}
\end{align}

\paragraph{Classification.} In node classification setups\footnote[3]{For graph classification, $Z_i$ and $y_i$ denote the `center' of the graph $i$ and its true class. We take the arithmetic mean of node embeddings in $\spd_n$ to produce $Z_i \in \spd_n$.}, we are given $\{Z_i, y_i\}_{i=1}^N$ on a dataset, with $N$ as the number of instances, $Z_i \in \spd_n$ as the $i$-th node embedding obtained from the final layer of a graph neural network, and $y_i \in \{1,\dots, K\}$ as the true class of $i$-th node. Let $h: \spd \mapsto \{1,\dots, K\}$ be a classifier that best predicts the label $y_i$ of a given input $Z_i$. Indeed, the input space of $h$ can be in various forms, not limited to $\spd_n$. Here we introduce three classifiers in two alternative input spaces: (a) $\RR^{d(d+1)/2}$ and (b) $S_n$\footnote{We also design several classifiers with the input space in $\spd_n$, but these do not yield better results than those in $S_n$.}. To do so, we first take Riemannian logarithm $\log \colon \spd_n \to S_n$ of each $Z_i$ at the identity. For (a), we vectorize the upper triangle elements of $X$ as $\mathbf{x} = (X_{1,1} \cdots X_{1,d}, X_{2,2}, \cdots, X_{d,d}) \in \RR^{d(d+1)/2}$, and then design two classifiers, i.e., \textsc{Linear-XE} (Linear Classifier coupled with Cross-Entropy loss) and \textsc{NC-MM} (Nearest Centroid Classifier with Multi-Margin loss). For (b), we design a SVM-like classifier \textsc{SVM-MM} acting in $S_n$, a similar approach to the proposal of Nguyen et al. \cite{nguyen2019spdHandGestureRecog}. We present the details of these classifiers in Appendix \ref{sec:classifiers}.

\section{Experiments}
In this section, we first perform experiments for node and graph classification, and then analyze the ability of three geometric spaces in arranging and separating nodes with different classes. Further, we compare the training efficiency of different spaces and the usefulness of three classifiers. Lastly, we discuss \textbf{product space} (the Cartesian product of Euclidean and hyperbolic spaces) and compare it with $\spd$ in Appendix \ref{sec:productspace}.

\label{sec:experiments}

\paragraph{Baselines.} To investigate the usefulness of different geometric spaces on graph neural networks, we choose five well-known graph architectures as representatives:
GCN \cite{kipf2017gnn}, GAT \cite{veličković2018graph}, Cheb \cite{defferrard2016convolutional}, SGC \cite{wu2019simplifying} and GIN \cite{xu2019gin}, and evaluate these architectures in Euclidean, hyperbolic and $\spd$ spaces. For the hyperbolic versions of these architectures, we use Poincar\'e models and extend the implementation of Poincar\'e GCN \cite{chami2019hgcnn} to other four architectures. 

\subsection{Node Classification}

\paragraph{Datasets.} 
We evaluate graph neural networks in the three spaces on 5 popular datasets for node classification: Disease \cite{anderson1992infectious}, Airport \cite{zhang2018link}, Pubmed \cite{namata2012query}, Citeseer and Cora \cite{sen2008collective}. Overall, each dataset has a single graph that contains up to thousands of labeled nodes. We use the public train, validation and test  splits of each dataset, and provide dataset statistics in Appendix \ref{sec:dataset_statistics}. Unlike previous works \cite{chami2019hgcnn,yu2022hyla}, we only use original node features to ensure a fair and transparent comparison.

\paragraph{Setup.} 
We compare three geometries, namely Euclidean, hyperbolic and $\spd$, in two low-dimensional spaces: (i) 6 dimensions: $\mathbb{R}^6$, $\mathbb{H}^6$ and $\spd_3$, and (ii) 15 dimensions: $\mathbb{R}^{15}$, $\mathbb{H}^{15}$ and $\spd_5$, a common choice of dimensions in previous work \cite{chami2019hgcnn,yu2022hyla}. 
The reason we considered for low-dimensional space is the following: If the structure of data matches the geometry of embedding space, a low-dimensional space can be leveraged efficiently for producing high-quality embeddings. If they do not match, a large dimension is needed to compensate for the wrong use of unsuitable geometric spaces. Here we investigate the efficiencies of different geometries in space use when given a small dimension.
We report mean accuracy and standard deviation of binary/multi-label classification results under 10 runs, and provide training details in Appendix \ref{sec:training_details}. 

\insertDatasetsA

\insertDimensions
\paragraph{Results.} 

Figure \ref{fig:datasetsA} shows the accuracy results of a node classification task in the three $6$-dimensional geometries across five datasets on five GNNs, see also Table \ref{tab:results-node-classification}. For graphs with $\delta$-hyperbolicity $>1$, $\spd_3$ achieves the best accuracy in all cases except the Cheb architecture on the Citeseer dataset. We also observe that the accuracy of $\spd$ is similar to hyperbolic space on the Airport dataset $\delta =1$. The Disease graph is a tree ($\delta=0$) and has optimal performance in hyperbolic space. The accuracy is much lower for these two tree-like datasets in Euclidean geometry for all GNNs except Cheb.

In the case of tree-like datasets, hyperbolic space provides not only accuracy for these tasks but also efficiency. Figure \ref{fig:6d_15d} compares $6$-dimensional hyperbolic space to $\RR^{15}$ and $\spd_5$ (also $15$-dimensional), showing that even a much smaller dimensional hyperbolic space achieves the best performance on Disease. Notably, the poor performance of Cheb across all spaces might be attributed to the low representational capacity of the first-order Chebyshev polynomial used in the graph neural network for embedding the tree structure of Disease. Results comparing the $6$d and $15$d geometries are reported in Table \ref{tab:dimensions-disease-cora} (appendix).

\subsection{Graph Classification}
\paragraph{Datasets.} We evaluate graph neural networks in three spaces on the popular TUDataset benchmark \cite{Morris:2020}. Here we focus on datasets with node features, and choose a sample of 4 popular datasets in two domains,
namely (a) Biology: ENZYMES \cite{schomburg2004brenda} and PROTEINS \cite{borgwardt2005protein}; (b) Molecules: COX2 \cite{sutherland2003spline} and AIDS \cite{riesen2008iam}.
Overall, each dataset instance has one labeled graph with dozens of nodes. We use the first split of train and test sets in the 10-fold cross-validation setup\footnote{Morris et al. \cite{Morris:2020} proposed to run 10-fold cross-validation by generating ten splits of train and test sets, and repeat this ten times to reduce model initialization.
This requires 100 runs for one setup, but it is impractical in our large-scale study.},
and select 10\% of the training set uniformly at random as the development set. We provide data statistics in Appendix \ref{sec:dataset_statistics}.

\paragraph{Setup.} Following Morris et al. \cite{Morris:2020}, we predict the class of an unlabeled graph by classifying its center. In particular, we produce the graph center by using mean pooling to take the arithmetic mean of node embeddings in a graph. 
To compare efficiency in space use, we conduct experiments in three spaces with the same dimension size of 36, namely $\mathbb{R}^{36}$, $\mathbb{H}^{36}$ and $\spd_8$, the smallest dimension size in the grid search from Morris et al. \cite{Morris:2020}. We report the mean accuracy and standard deviation of graph classification results under 10 runs, and provide training details in Appendix \ref{sec:training_details}. 

\paragraph{Results.} 

Figure \ref{fig:datasetsB} shows the accuracy results of a graph classification task in three geometries across five datasets on five GNNs, see also Table \ref{tab:results-graph-classification}. Figure \ref{fig:hyperbolicity} shows the distribution of $\delta$-hyperbolicity over instances. We see that $\spd_8$ achieves better or similar accuracy than its counterparts of the same dimenison in all cases except Cheb on COX2 and GIN on ENZYMES. On the AIDS dataset, $\spd$ achieves much better accuracy across all GNNs, and on ENZYMES $\spd$ achieves much better accuracy on SGC.

We also observe that hyperbolic space does not yield much increased accuracy over Euclidean space in most cases, except the AIDS data set. Furthermore, $\spd$ significantly outperforms hyperbolic space on the AIDS dataset but not the COX2 data set. Both of these datasets have the property that almost all instances are tree-like ($\delta \le 1$) (see Figure \ref{fig:hyperbolicity}), but the hyperbolicity constants are less concentrated in the AIDS dataset than in the COX2 dataset. It is possible that the flexibility of $\spd$ explains the increased performance over hyperbolic space in this case. For example, $\spd$ admits totally geodesic submanifolds isometric to hyperbolic spaces of varying constant curvatures. It would be interesting to find out, for example, if these graphs of different hyperbolicity constants stay near copies of hyperbolic space of different curvatures.
\insertDatasetsB

\insertHyperbolicity
\subsection{Analysis}
\label{sec:analysis-case}

\paragraph{Class Separation.} 

Figure \ref{fig:class_separation} shows a visualization of  node embeddings obtained from the final layer of SGConv on Cora. In each space, we vectorize node embeddings, and then use PCA to extract the top 3 dimensions. We then look at the projections to that 3-dimensional space by further projecting to the x-y, y-z, and x-z planes. For instance, the x-y plane is the projection to the top 2 dimensions of PCA. The x-x, y-y and z-z planes show the informativeness of each dimension in terms of class separation. 

Figure \ref{fig:class_separation} (a) depicts the case when the ambient geometry is Euclidean. In this example, nodes from the pink, blue and green classes are well-separated but the nodes in red and orange cannot be easily distinguished. Figure \ref{fig:class_separation} (b) depicts the case when the nodes are embedded in the Poincar\'e ball model of hyperbolic space. In the x-z plane five classes are well-separated, including the red and orange classes. Figure \ref{fig:class_separation} (c) depicts the case when the nodes are embedded into $\spd$ where the best class separation is achieved. Indeed, the Cora graph has hyperbolicity constant $\delta=11$, so one cannot expect it to embed well into hyperbolic space.

\insertClassSeparation

\paragraph{Training Time.} 

As a case study on the impact of the choice of the latent space geometry on the training time of GNN models, Figure \ref{fig:runningtime} compares the training efficiency of three graph neural networks on Citeseer across three 6-dimensional manifolds: $\mathbb{R}^6$, $\mathbb{H}^6$, and $\spd_3$.
Overall, we see that models with Euclidean latent spaces needed the least amount of training time, but produced the lowest accuracy. Moreover, hyperbolic space slows down the training in Euclidean space by up to four times, and the effect on accuracy is either slightly positive or negative. For instance, using hyperbolic space with the SGC architecture only brought small accuracy improvements, while applying it to GCN resulted in a drop in accuracy. On the other hand, even though $\spd$ models require nearly double the training time of hyperbolic models, the $\spd$ models bring big improvements in accuracy, not only on Citeseer in the current study case, but also on many datasets such as AIDS and PROTEINS (see Figure \ref{fig:datasetsB}). We note that the longer training times for $\spd$ models can be attributed to the involvement of eigendecompositions. However, the benefits of using $\spd$ space in graph neural networks appear to outweigh this drawback.

\insertRunningtime

\paragraph{Classifiers.} 

Our three classification layers are built upon traditional classification methods. \textsc{LINEAR-XE} and \textsc{SVM-MM} are both linear classifiers that separate classes with hyperplanes, differing in the choice of loss functions: cross-entropy and multi-margin loss. In contrast, \textsc{NC-MM} layer learns class-specific centroids and then determining the class of an unlabeled node (or graph) by examining which centroid it is closest to according to a similarity function.

Table \ref{tab:classifiers} shows the usefulness of our classifiers in $\spd$ on Citesser and Cora. Overall, we see that the MM-based classifiers (\textsc{SVM-MM} and \textsc{NC-MM}) are often helpful, outperforming \textsc{Linear-XE} in $\spd$, and when they succeed, their improvements are substantial. 
This means the benefits of using $\spd$ and these advanced classifiers are complementary, resulting in stacked performance gains. This is an important finding as it hints at the possibility of accommodating more advanced classification methods recently developed in Euclidean space, when constructing graph neural networks in $\spd$. Note that the results on other datasets are similar, which we present in Appendix \ref{sec:results_node} and \ref{sec:results_graph}. 

\insertTableClassifiers

\section{Conclusions}

This work brings sophisticated
geometric tools to graph neural networks (GNNs).  
Following the maxim ‘complex data requires complex geometry’, we leverage the flexibility of the space of symmetric positve definite ($\spd$) matrices to construct GNNs which do not require 
careful prior knowledge of graph topologies.  
This is a distinct advantage over familiar spaces such as Euclidean, spherical or hyperbolic geometries, where only
narrow classes of graphs embed with low distortion.

To operate GNNs in $\spd$, we designed several building blocks, and developed a library (SPD4GNN) that enables training five popular GNNs in $\spd$, Euclidean and hyperbolic spaces. 
Our results confirm the strong connection between graph topology and embedding geometry: GNNs in $\spd$ provide big improvements on graph datasets with multi-modal structures, with their counterparts in hyperbolic space performing better on strictly tree-like graphs.

Determining the optimal classifier for training GNNs in the complex geometry of $\spd$ is challenging, and presents an avenue for continued improvement.  This work only begins the process of designing geometrically meaningful classifiers and identifying the conditions which guarantee good performance.  
Additional performance gains may come through careful implementation of the computationally demanding functions in $\spd$.  
While this work contains techniques for accelerating computations in $\spd$, further optimization is likely possible.

Constructing tools to aid the interpretability of $\spd$ embeddings is an important direction of future work, including quantitative measures for (a) comparing the geometry of the learned embeddings to the real-world graphs' topology and (b) understanding how the geometric features of $\spd$ are leveraged in for graph tasks.
While the results of this current work suggest some of SPD’s superior performance may be due to graphs of varying hyperbolicity finding geometric subspaces optimally adapted to their curvature,
such measures would enable the precise quantitative analysis required for verification. 

\section{Ethical Considerations}
We have not identified any immediate ethical concerns such as bias and discrimination, misinformation dissemination, privacy issues, originating from the contributions presented in this work. However, it is important to note that our $\spd$ models use computationally demanding functions, such as determining eigenvalues and eigenvectors, which may incur a negative environmental impact due to increased energy consumption. Nevertheless, SPD models do not outsuffer Euclidean and hyperbolic counterparts in terms of computational overhead. This is because Euclidean and hyperbolic models would require substantial computing resources when dealing with larger dimensions, a necessity for compensating for the challenges of embedding complex graphs into these ill-suited spaces.

\section*{Acknowledgements}

We thank Anna Wienhard and Maria Beatrice Pozetti for insightful discussions, as well as the anonymous reviewers for their
thoughtful feedback that greatly improved the
texts. This work has been supported by the Klaus Tschira Foundation, Heidelberg, Germany, as well as under Germany’s Excellence Strategy EXC-2181/1 - 390900948 (the Heidelberg STRUCTURES Cluster of Excellence). 

\bibliography{main}

\begin{thebibliography}{10}
\providecommand{\url}[1]{\texttt{#1}}
\providecommand{\urlprefix}{URL }
\providecommand{\doi}[1]{https://doi.org/#1}

\bibitem{anderson1992infectious}
Anderson, R.M., May, R.M.: Infectious diseases of humans: dynamics and control.
  Oxford university press (1992)

\bibitem{barcelo2022weisfeiler}
Barcelo, P., Galkin, M., Morris, C., Orth, M.R.: Weisfeiler and leman go
  relational. In: The First Learning on Graphs Conference (2022),
  \url{https://openreview.net/forum?id=wY_IYhh6pqj}

\bibitem{borgwardt2005protein}
Borgwardt, K.M., Ong, C.S., Sch{\"o}nauer, S., Vishwanathan, S., Smola, A.J.,
  Kriegel, H.P.: Protein function prediction via graph kernels. Bioinformatics
  \textbf{21}(suppl\_1),  i47--i56 (2005)

\bibitem{bronstein2017geometric}
Bronstein, M.M., Bruna, J., LeCun, Y., Szlam, A., Vandergheynst, P.: Geometric
  deep learning: going beyond euclidean data. IEEE Signal Processing Magazine
  \textbf{34}(4),  18--42 (2017)

\bibitem{brooks2019riemBNforSPD}
Brooks, D., Schwander, O., Barbaresco, F., Schneider, J.Y., Cord, M.:
  Riemannian batch normalization for {SPD} neural networks. In: Wallach, H.,
  Larochelle, H., Beygelzimer, A., d\textquotesingle Alch\'{e}-Buc, F., Fox,
  E., Garnett, R. (eds.) Advances in Neural Information Processing Systems.
  vol.~32, pp. 15489--15500. Curran Associates, Inc. (2019),
  \url{https://proceedings.neurips.cc/paper/2019/file/6e69ebbfad976d4637bb4b39de261bf7-Paper.pdf}

\bibitem{brooks2019riemannianRadarData}
Brooks, D.A., Schwander, O., Barbaresco, F., Schneider, J.Y., Cord, M.:
  Exploring complex time-series representations for riemannian machine learning
  of radar data. In: ICASSP 2019 - 2019 IEEE International Conference on
  Acoustics, Speech and Signal Processing (ICASSP). pp. 3672--3676 (2019).
  \doi{10.1109/ICASSP.2019.8683056}

\bibitem{buyalo2005embedding}
Buyalo, S., Schroeder, V.: Embedding of hyperbolic spaces in the product of
  trees. Geometriae Dedicata  \textbf{113}(1),  75--93 (2005)

\bibitem{cannon1997hyperbolic}
Cannon, J.W., Floyd, W.J., Kenyon, R., Parry, W.R., et~al.: Hyperbolic
  geometry. Flavors of geometry  \textbf{31}(59-115), ~2 (1997)

\bibitem{chakraborty2020manifoldNet}
Chakraborty, R., Bouza, J., Manton, J., Vemuri, B.C.: Manifoldnet: A deep
  neural network for manifold-valued data with applications. IEEE Transactions
  on Pattern Analysis and Machine Intelligence pp.~1--1 (2020).
  \doi{10.1109/TPAMI.2020.3003846}

\bibitem{chakraborty2018recurrentSPD}
Chakraborty, R., Yang, C.H., Zhen, X., Banerjee, M., Archer, D., Vaillancourt,
  D., Singh, V., Vemuri, B.: A statistical recurrent model on the manifold of
  symmetric positive definite matrices. In: Bengio, S., Wallach, H.,
  Larochelle, H., Grauman, K., Cesa-Bianchi, N., Garnett, R. (eds.) Advances in
  Neural Information Processing Systems. vol.~31. Curran Associates, Inc.
  (2018),
  \url{https://proceedings.neurips.cc/paper/2018/file/7070f9088e456682f0f84f815ebda761-Paper.pdf}

\bibitem{chami2019hgcnn}
Chami, I., Ying, Z., R\'{e}, C., Leskovec, J.: Hyperbolic graph convolutional
  neural networks. In: Advances in Neural Information Processing Systems 32,
  pp. 4869--4880. Curran Associates, Inc. (2019),
  \url{https://proceedings.neurips.cc/paper/2019/file/0415740eaa4d9decbc8da001d3fd805f-Paper.pdf}

\bibitem{defferrard2016convolutional}
Defferrard, M., Bresson, X., Vandergheynst, P.: Convolutional neural networks
  on graphs with fast localized spectral filtering. Advances in neural
  information processing systems  \textbf{29} (2016)

\bibitem{Defferrard2020DeepSphere}
Defferrard, M., Milani, M., Gusset, F., Perraudin, N.: Deep{S}phere: A
  graph-based spherical {CNN}. In: International Conference on Learning
  Representations (2020), \url{https://openreview.net/forum?id=B1e3OlStPB}

\bibitem{di2022graph}
Di~Giovanni, F., Rowbottom, J., Chamberlain, B.P., Markovich, T., Bronstein,
  M.M.: Graph neural networks as gradient flows. arXiv preprint
  arXiv:2206.10991  (2022)

\bibitem{dong2017spdToFaceRecog}
Dong, Z., Jia, S., Zhang, C., Pei, M., Wu, Y.: Deep manifold learning of
  symmetric positive definite matrices with application to face recognition.
  In: Proceedings of the Thirty-First AAAI Conference on Artificial
  Intelligence. p. 4009–4015. AAAI'17, AAAI Press (2017)

\bibitem{eliasof2021pde}
Eliasof, M., Haber, E., Treister, E.: Pde-gcn: Novel architectures for graph
  neural networks motivated by partial differential equations. Advances in
  Neural Information Processing Systems  \textbf{34},  3836--3849 (2021)

\bibitem{gao2019robustRepreSPD}
Gao, Z., Wu, Y., Bu, X., Yu, T., Yuan, J., Jia, Y.: Learning a robust
  representation via a deep network on symmetric positive definite manifolds.
  Pattern Recognition  \textbf{92},  1--12 (2019).
  \doi{https://doi.org/10.1016/j.patcog.2019.03.007},
  \url{https://www.sciencedirect.com/science/article/pii/S0031320319301062}

\bibitem{gu2019lmixedCurvature}
Gu, A., Sala, F., Gunel, B., Ré, C.: Learning mixed-curvature representations
  in product spaces. In: International Conference on Learning Representations
  (2019), \url{https://openreview.net/forum?id=HJxeWnCcF7}

\bibitem{gu2018learning}
Gu, A., Sala, F., Gunel, B., Ré, C.: Learning mixed-curvature representations
  in product spaces. In: International Conference on Learning Representations
  (2019), \url{https://openreview.net/forum?id=HJxeWnCcF7}

\bibitem{sage:2017}
Hamilton, W., Ying, Z., Leskovec, J.: Inductive representation learning on
  large graphs. In: Guyon, I., Luxburg, U.V., Bengio, S., Wallach, H., Fergus,
  R., Vishwanathan, S., Garnett, R. (eds.) Advances in Neural Information
  Processing Systems. vol.~30. Curran Associates, Inc. (2017),
  \url{https://proceedings.neurips.cc/paper/2017/file/5dd9db5e033da9c6fb5ba83c7a7ebea9-Paper.pdf}

\bibitem{helgason1078diffGeom}
Helgason, S.: Differential geometry, Lie groups, and symmetric spaces. Academic
  Press New York (1978)

\bibitem{huang2017riemannianNetForSPDMatrix}
Huang, Z., Van~Gool, L.: A {R}iemannian network for {SPD} matrix learning. In:
  Proceedings of the Thirty-First AAAI Conference on Artificial Intelligence.
  p. 2036–2042. AAAI'17, AAAI Press (2017)

\bibitem{kingma2014Adam}
Kingma, D.P., Ba, J.: Adam: A method for stochastic optimization. International
  Conference on Learning Representations  \textbf{abs/1412.6980} (2014)

\bibitem{kipf2017gnn}
Kipf, T.N., Welling, M.: Semi-supervised classification with graph
  convolutional networks. In: 5th International Conference on Learning
  Representations, {ICLR} 2017, Toulon, France, April 24-26, 2017, Conference
  Track Proceedings (2017), \url{https://openreview.net/forum?id=SJU4ayYgl}

\bibitem{krioukov2010hyperbolic}
Krioukov, D., Papadopoulos, F., Kitsak, M., Vahdat, A., Bogun{\'a}, M.:
  Hyperbolic geometry of complex networks. Physical Review E  \textbf{82}(3),
  036106 (2010)

\bibitem{hgnn:2019}
Liu, Q., Nickel, M., Kiela, D.: Hyperbolic graph neural networks. In: Wallach,
  H., Larochelle, H., Beygelzimer, A., d\textquotesingle Alch\'{e}-Buc, F.,
  Fox, E., Garnett, R. (eds.) Advances in Neural Information Processing
  Systems. vol.~32. Curran Associates, Inc. (2019)

\bibitem{lopez2021symmetric}
L\'opez, F., Pozzetti, B., Trettel, S., Strube, M., Wienhard, A.: Symmetric
  spaces for graph embeddings: A finsler-riemannian approach. In: Meila, M.,
  Zhang, T. (eds.) Proceedings of the 38th International Conference on Machine
  Learning. Proceedings of Machine Learning Research, vol.~139, pp. 7090--7101.
  PMLR (18--24 Jul 2021), \url{http://proceedings.mlr.press/v139/lopez21a.html}

\bibitem{lopez2021gyroSPD}
L\'opez, F., Pozzetti, B., Trettel, S., Strube, M., Wienhard, A.: Vector-valued
  distance and gyrocalculus on the space of symmetric positive definite
  matrices. In: Larochelle, H., Ranzato, M., Hadsell, R., Balcan, M.F., Lin, H.
  (eds.) Advances in Neural Information Processing Systems. vol.~34. Curran
  Associates, Inc. (2021)

\bibitem{Morris:2020}
Morris, C., Kriege, N.M., Bause, F., Kersting, K., Mutzel, P., Neumann, M.:
  Tudataset: A collection of benchmark datasets for learning with graphs. In:
  ICML 2020 Workshop on Graph Representation Learning and Beyond (GRL+ 2020)
  (2020), \url{www.graphlearning.io}

\bibitem{namata2012query}
Namata, G., London, B., Getoor, L., Huang, B., Edu, U.: Query-driven active
  surveying for collective classification. In: 10th International Workshop on
  Mining and Learning with Graphs. vol.~8, p.~1 (2012)

\bibitem{nguyen2019spdHandGestureRecog}
Nguyen, X.S., Brun, L., Lezoray, O., Bougleux, S.: A neural network based on
  {SPD} manifold learning for skeleton-based hand gesture recognition. In:
  Proceedings of the IEEE/CVF Conference on Computer Vision and Pattern
  Recognition (CVPR) (June 2019)

\bibitem{riesen2008iam}
Riesen, K., Bunke, H.: Iam graph database repository for graph based pattern
  recognition and machine learning. In: Joint IAPR International Workshops on
  Statistical Techniques in Pattern Recognition (SPR) and Structural and
  Syntactic Pattern Recognition (SSPR). pp. 287--297. Springer (2008)

\bibitem{rusch2022graph}
Rusch, T.K., Chamberlain, B., Rowbottom, J., Mishra, S., Bronstein, M.:
  Graph-coupled oscillator networks. In: International Conference on Machine
  Learning. pp. 18888--18909. PMLR (2022)

\bibitem{satorras2021n}
Satorras, V.G., Hoogeboom, E., Welling, M.: E (n) equivariant graph neural
  networks. In: International conference on machine learning. pp. 9323--9332.
  PMLR (2021)

\bibitem{schomburg2004brenda}
Schomburg, I., Chang, A., Ebeling, C., Gremse, M., Heldt, C., Huhn, G.,
  Schomburg, D.: Brenda, the enzyme database: updates and major new
  developments. Nucleic acids research  \textbf{32}(suppl\_1),  D431--D433
  (2004)

\bibitem{sen2008collective}
Sen, P., Namata, G., Bilgic, M., Getoor, L., Galligher, B., Eliassi-Rad, T.:
  Collective classification in network data. AI magazine  \textbf{29}(3),
  93--93 (2008)

\bibitem{sonthalia2020tree}
Sonthalia, R., Gilbert, A.: Tree! i am no tree! i am a low dimensional
  hyperbolic embedding. Advances in Neural Information Processing Systems
  \textbf{33},  845--856 (2020)

\bibitem{sutherland2003spline}
Sutherland, J.J., O'brien, L.A., Weaver, D.F.: Spline-fitting with a genetic
  algorithm: A method for developing classification structure- activity
  relationships. Journal of chemical information and computer sciences
  \textbf{43}(6),  1906--1915 (2003)

\bibitem{veličković2018graph}
Veličković, P., Cucurull, G., Casanova, A., Romero, A., Liò, P., Bengio, Y.:
  Graph attention networks. In: International Conference on Learning
  Representations (2018), \url{https://openreview.net/forum?id=rJXMpikCZ}

\bibitem{wu2019simplifying}
Wu, F., Souza, A., Zhang, T., Fifty, C., Yu, T., Weinberger, K.: Simplifying
  graph convolutional networks. In: International conference on machine
  learning. pp. 6861--6871. PMLR (2019)

\bibitem{xu2019gin}
Xu, K., Hu, W., Leskovec, J., Jegelka, S.: How powerful are graph neural
  networks? In: 7th International Conference on Learning Representations,
  {ICLR} 2019, New Orleans, LA, USA, May 6-9, 2019. OpenReview.net (2019),
  \url{https://openreview.net/forum?id=ryGs6iA5Km}

\bibitem{yu2022hyla}
Yu, T., De~Sa, C.: Hyla: Hyperbolic laplacian features for graph learning.
  arXiv preprint arXiv:2202.06854  (2022)

\bibitem{zhang2018link}
Zhang, M., Chen, Y.: Link prediction based on graph neural networks. Advances
  in neural information processing systems  \textbf{31} (2018)

\bibitem{zhang2018manifoldToManifold}
Zhang, T., Zheng, W., Cui, Z., Li, C.: Deep manifold-to-manifold transforming
  network. In: 2018 25th IEEE International Conference on Image Processing
  (ICIP). pp. 4098--4102 (2018). \doi{10.1109/ICIP.2018.8451626}

\end{thebibliography}
\bibliographystyle{splncs04}

\newpage
\appendix

\section{Training Details}
\label{sec:training_details}

For both node and graph classification, we use grid search to tune the same set of hyperparameters for each graph architecture on the development set of a given dataset, and repeat the tuning process three times over Euclidean and hyperbolic spaces, the product of thereof, and the $\spd$ space. We consider 5 hyperparameters: (a) learning rate $\in \{0.1, 0.01, 0.001\}$; (b) dropout $\in \{0, 0.5\}$; (c) weight-decay $\in \{0, 0.005, 0.0005\}$ used to regularize a model by shrinking the L2 norm of model weights; (d) nonlinearity $\in \{\rm{TgReEig}, \rm{ReEig}\}$ only used for the $\spd$ space\footnote{In Euclidean and hyperbolic cases, we apply ReLU to Euclidean space and to the tangent space of hyperbolic manifold, as suggested by Chami et al. \cite{chami2019hgcnn}.} and (e) $C$ $\in \{0.5, 0.05, 0.005, 0.0005\}$ used to control the size of SVM hyperplane. For node classification, we follow Chami et al. \cite{chami2019hgcnn} and set batch size to a total number of graph nodes in a dataset. We train individual graph architectures for a maximum of 500 epochs, and stop training when the loss on the development set has not decreased for 200 epochs. Regarding graph classification, We set batch size to 32, and set the epochs to a maximum of 200 for training, accompanied by a patience of 100 epochs for early stopping. In all setups, we use a stack of two graph layers that perform message passing twice at each iteration. 

For optimization, considering the model parameters of $\spd$ graph neural networks are in Euclidean space, we use Adam optimizer \cite{kingma2014Adam} to tune these parameters, and minimize (a) the cross-entropy loss for linear classifier and (b) the multi-margin hinge loss for both SVM and centriod-based classifiers. Notably, our graph neural networks in $\spd$ take longer time to train than the Euclidean and hyperbolic counterparts due to the requirement of computing costly eigenvalues and eigenvectors at each iteration. 
We choose Euclidean over Riemannian optimization, as the latter cannot guarantee the symmetry positive definiteness in $\spd$ matrices, a potential risk incurring numerical instability.

\section{Graph Architectures}
\label{sec:graph_architecture}
Unlike GCN, GAT and Cheb, SGC and GIN both compute propagation ahead of applying feature transformation. We contrast their architectures in different spaces in Table \ref{tab:architectures2}.

\insertSGCGIN

\section{Dataset Statistics}
Table \ref{tab:nc-dataset} and \ref{tab:gc-dataset} present data statistics for the benchmark datasets in node and graph classification setups.

\label{sec:dataset_statistics}

\begin{table}[t]
\caption{
    Dataset statistics for node classification.}
    \footnotesize
    \centering
    \begin{tabular}{lccccc}
    \toprule
    Dataset & \# Nodes & \# Edges& \# Classes & \# Features \\
    \midrule
    Cora & 2,708 & 5,429 & 7 & 1,433 \\
    Citeseer & 3,327 & 4,732 & 6 & 3,703 \\
    Pubmed &  19,717  & 44,338 & 3 & 500 \\
    Disease & 1,044 & 1,043 & 2 & 1,000 \\
    Airport & 3,188 & 18,631 & 4 & 4 \\
    \bottomrule
    \end{tabular}

    \label{tab:nc-dataset}
\end{table} 

\begin{table}[t]
\caption{
    Dataset statistics for graph classification. \# Nodes  and \# Edges denote the number of nodes and edges on average across graphs.}
    \scriptsize
    \centering
    \begin{tabular}{lccccc}
    \toprule
    Dataset & \# Graphs & \# Nodes & \# Edges& \# Classes & \# Features \\
    \midrule
    ENZYMES & 600 & 32.6 & 124.3 & 6 & 3 \\
    PROTEINS & 1,113 & 39.1 & 145.6 & 2 & 3 \\
    COX2 & 467 & 41.2 & 43.4 & 2 &  35 \\
    AIDS & 2,000 & 15.6 & 16.2 & 2 & 38 \\
    \bottomrule
    \end{tabular}

    \label{tab:gc-dataset}
\end{table} 

\section{Results in Node Classification}
\label{sec:results_node}
\insertTableNodeClassification
Table \ref{tab:results-node-classification} shows a  detailed breakdown of results for node classification. 
\insertTableDimensionsDiseaseCora

\section{Results in Graph Classification}
\label{sec:results_graph}
\insertTableGraphClassification
Table \ref{tab:results-graph-classification} shows a  detailed breakdown of results for graph classification. 

\section{Hitchhiker's Guide to Numerical Problems}
\label{sec:numerical_issues}

Unlike Euclidean space, computing work in $\spd$ does not utilize matrix multiplication, but rather relies on the use of gyrocalculus and algebraic operations in symmetric space. However, these operations may incur numerical instability. In what follows we present a list of numerical issues identified in this work, and provide solutions to address them.

\paragraph{Repeated Eigenvalues.}GNNs in $\spd$ extensively utilize logmap and expmap to switch the space of node embeddings between $\spd$ and the space of symmetric matrices.  
Doing this requires the determination of 
the orthogonal basis and eigenvalues of an $\spd$ matrix (or a symmetric matrix). However, when eigenvalues are not unique, the orthogonal basis is not well-defined, i.e., any vector in the eigenspace associated to a repeated eigenvalue is a legitimate eigenvector. This could incur the reproducibility issue as an orthogonal basis could be determined differently across machines, libraries, and even multiple runs. Further, the gradients of eigenvectors in the degenerate case are not well-supported in popular libraries such as Pytorch---where the gradients are stable only when the corresponding eigenvalues are unique. We address this issue by adding a random noise following a normal distribution with zero mean and the standard deviation of 0.001 to both $\spd$ and symmetric matrices.

\paragraph{Non-positive definiteness of Node Embeddings.} Machinery computing typically involves the use of imprecise floating-point systems to compute and represent numbers, particularly efficient to compute but prone to round-off and truncation errors, such as representing an irrational number approximately with a finite amount of decimals. 
These errors could affect the positive-definiteness of node embeddings by yielding symmetric matrices with slightly negative eigenvalues. When node embeddings are no longer in $\spd$, 
a great number of algebraic $\spd$\ operations becomes numerical unstable, such as computing the geodesic distance between two points in $\spd$\footnote{For instance, computing geodesic distance requires taking the square root of eigenvalues. In this case, numerical instability arises as eigenvalues are negative.}, mapping points from $\spd$ to the tangent space at a given point, computing gyro-addition. To this end, we take a two-step process: (a) enabling double precision floating-point format in computing, and (b) clamping eigenvalues to ensure them being positive. 
    
\paragraph{Library Issue.} 
Popular libraries such as Pytorch have struggled to efficiently compute eigenvectors and eigenvalues for long. 
We benchmarked the running time of two factorization algorithms in Pytorch: (a) finding the eigenvectors and eigenvalues via eigenvalue decomposition and (b) finding the singular vectors and values via singular value decomposition (SVD) of 5,000 symmetric matrices with the size of $8\times 8$. We found that eigenvalue decomposition was 15 times slower to run than SVD.
For that reason, we utilize SVD coupled with a sign correction to compute eigenvectors and eigenvalues.
We found that our solution is faster than the Pytorch in-house eigenvalue decomposition by approximately 12 times. 
In June 2022, Pytorch released an update that finally improved the speed of computing eigenvectors and eigenvalues. 

\section{Classifiers}
\label{sec:classifiers}

\paragraph{Linear-XE.} First, we consider a linear classification layer. The layer maps an input SPD matrix $\mathbf{Z}$ to its logarithm $\mathbf{X} = \log(\mathbf{Z})$, vectorizes the upper triangle elements of $\mathbf{X}$ as $\mathbf{x} = (X_{1,1} \cdots X_{1,d}, X_{2,2}, \cdots, X_{d,d}) \in \RR^{d(d+1)/2}$, then applies a linear transformation $\mathbf{y} = \mathbf{W} \mathbf{x}$, where $\mathbf{W} \in \RR^{K \times d(d+1)/2}$ is a learnable parameter matrix, and $K$ is the number of classes. The model parameters are then trained with a standard cross entropy loss objective applied to $\mathbf{y}$. Finally, predictions are made after training according to $\argmax_k y_k$ with $y_k$ as the $k$th component of $\mathbf{y}$.

\paragraph{SVM-MM.}
The second classification layer we consider is a modification of a standard SVM approach. As before in the \textbf{Linear-XE} case, we begin by mapping an input $\spd$ matrix $\mathbf{Z}$ to its logarithim $\mathbf{X} = \log (\mathbf{Z})$. Instead of vectorizing $\mathbf{Z}$ and learning a linear transformation, in this case our implementations learns linear functionals applied directly to $\mathbf{X}$. This is done by identifying $S_n$ with its dual space $S_n^\ast$ via $W \mapsto (X \mapsto Tr(WX))$. The loss function then consists of a standard multi-class hinge loss function and a regularization term, analogous to a term proportional to $\lambda |W|^2$ in Euclidean space. Specifically, let $C = \frac{1}{N} \sum_{n=i}^N X_i$, the arithmetic mean of the subset $\{X_i\}_{i=1}^N$ in $\spd_n$, which is again in $\spd_n$. Then the loss function we propose is
\begin{align*}
    & \mathrm{Loss}_{\mathbf{SVM-MM}}(X_1,\dots,X_N;W_1,\dots,W_K) \\
    & = \lambda \sum_{k=1}^K Tr(W_kCW_kC) +\\ 
    & \frac{1}{NK^2} \sum_{i=1}^N \sum_{k=1}^K \sum_{\substack{l=1 \\ l\ne k}}^K \max(0,1- Tr(W_kX_i) + Tr(W_l X_i)) 
\end{align*}
where $\lambda$ is a hyperparameter. Predictions are made after training according to $\argmax_k (X \mapsto Tr(W_kX))$.

\paragraph{NC-MM.} Lastly, we consider a centroid-based layer for classifying $K$ classes. The layer is parametrized by $K$ centroids $\{\mu_k\}_{k=1}^K \subset \RR^{d(d+1)/2}$, $K$ SPD matrices $\{P_k\}_{k=1}^K \subset \spd(d(d+1)/2,\RR)$, and $K$ bias terms $\{b_k\}_{k=1}^K \subset \RR$. Given an input SPD matrix $Z$, the layer first maps $Z$ to its logarithm $X = \log(Z)$, and then vectorizes the upper triangle elements of $X$ as $x = (X_{1,1} \cdots X_{1,d}, X_{2,2}, \cdots, X_{d,d}) \in \RR^{d(d+1)/2}$. A centroid-based similarity value for each class $k$ is then computed as $\similar(x, \mu_k) := -\frac{1}{2} (x - \mu_k) P_k (x - \mu_k)^\top + b_k$. The model parameters are trained using a standard multi-hinge loss objective applied to the feature vector $\left(\similar(x, \mu_k)\right)_{k=1}^K$. Finally, predictions are made after training according to $\argmax_k \similar(x, \mu_k)$.

\section{Product Manifold}
\label{sec:productspace}

Let $M_1, M_, \dots, M_k$ be a sequence of smooth manifolds. The product manifold is given by the Cartesian product $M = M_1 \times M_2 \times \dots \times M_k$. Each point $p \in M$ has the coordinates $p = (p_1, \dots, p_k)$, with $p_i \in M_i$ for all $i$. Similarly, a tangent vector $v \in T_p M$ can be written as $(v_1, \dots , v_k)$, with each $v_i \in T_{p_i}M_i$. If each $M_i$ is equipped with a Riemannian metric $g_i$, then the product manifold $M$ can be given the product metric where $g(v,w) = \sum_{i=1}^k g_i(v_i,w_i)$.

\paragraph{Application to Graph Neural Networks.} We construct graph convolutional network (GCN) in product manifold given by $M = \mathbb{H}^m \times \mathbb{R}^m$. Let $Z_{i}^{l}=(Z_{i,1}^l, Z_{i,2}^l) \in M$ be the embedding of the node $i$ at the $l$ layer, with $Z_{i,1}^l \in \mathbb{H}^m$ and $Z_{i,2}^l \in \mathbb{R}^m$. In the following, we implement three building blocks that enable GCN to operate in product manifold.

\begin{itemize}
    \item \textit{Feature transformation} combines hyperbolic and linear maps to transform each point $Z_{i}^{l-1}$ in $M$  at each layer $l$, denoted by:

    \begin{equation*}
    Q_i^l=W^{l} \otimes Z_{i}^{l-1} = 
    \begin{pmatrix}
    W_{11}^{l} & W_{12}^{l}\\
    W_{21}^{l} & W_{22}^{l}
    \end{pmatrix}
    \otimes
    \begin{pmatrix}
         Z_{i,1}^{l-1}\\
         Z_{i,2}^{l-1}
    \end{pmatrix}
    = 
    \begin{pmatrix}
         (W_{11}^{l}\odot Z_{i,1}^{l-1}) \oplus (W_{12}^{l}\odot \exp(Z_{i,2}^{l-1}))\\
         W_{21}^{l}\log(Z_{i,1}^{l-1})+W_{22}^{l} Z_{i,2}^{l-1}
    \end{pmatrix}
    \end{equation*}
    Where $W_{ij}^{l} \in GL(m, \mathbb{R}) 
    $ for all $i,j$ (with learnable weights adjusted to match training data), $Q_i^l = (Q_{i,1}^l, Q_{i,2}^l) \in \mathbb{H}^{m} \times \mathbb{R}^{m}$, $\oplus$ is M\"obius addition, $\exp$ and $\log$ are the mappings between hyperbolic space and its tangent space, and $\odot$ is hyperbolic matrix multiplication, denoted by $W_{11}^{l}\odot Z_{i,1}^{l-1}:= \exp(W_{11}^{l} \log(Z_{i,1}^{l-1}))$ for example.

    \item \textit{Propagation} aggregates information from all the neighbors $\mathcal{N}(i)$ of a given node $i$, with $k_{i,j}$ as the weight for node $i$ and $j$:
    $$P_{i}^{l} = (\exp(\sum_{j \in \mathcal{N}(i)} k_{i, j} \log(Q_{j,1}^l)), \sum_{j \in \mathcal{N}(i)} k_{i, j} Q_{j,2}^l))$$
    Where $P_i^l = (P_{i,1}^l, P_{i,2}^l) \in \mathbb{H}^m \times \mathbb{R}^m$, and $k_{i,j} = c_i^{-\frac{1}{2}} c_j^{-\frac{1}{2}}$ with $c_i$ as the cardinality of $\neighs(i)$.

    \item \textit{Bias and Non-linearity} adds the biases to the propagation results followed by applying two non-linear functions:
    
    $$Z_{i}^{l} = (\varphi_1 (P_{i,1}^{l} \oplus B_{1}^{l}), \varphi_2 (P_{i,2}^{l} + B_{2}^{l}))$$
    
    Where $\varphi_1: x \mapsto \exp(\mathrm{ReLU} (\log(x))$, $\varphi_2: x \mapsto \mathrm{ReLU} (x)$, $\oplus$ is M\"obius addition, and $B_{1}^{l} \in \mathbb{H}^m$ and $B_{2}^{l} \in \mathbb{R}^m$ are learnable bias weights.
    
\end{itemize}

\paragraph{Results.} 

Table \ref{tab:product-space} compares product space with $\spd$, Euclidean and hyperbolic spaces.
Results show that the product space $\mathbb{H}^3 \times \mathbb{R}^3$ considerably underperforms $\spd_3$, despite both combining Euclidean and hyperbolic subspaces. Furthermore, the product space performs worse than the Euclidean and hyperbolic counterparts ($\mathbb{R}^6$ and $\mathbb{H}^6$) on non-hyperbolic datasets including Pubmed, Citeseer and Cora. 

This matches the mathematical expectation that $\spd$, which combines Euclidean and hyperbolic features in a more intricate way than product manifolds, has a higher representation capacity, even though both classes of spaces exhibit mixed curvature. For that reason, $\spd$ more readily accommodates a wider variety of graph data.

\begin{table}
\centering
\caption{Comparison to the product space ($\mathbb{H}^3 \times \mathbb{R}^3$) for node classification. Results are from GCNConv coupled with Linear-XE.}
\scriptsize
\begin{tabular}{c |l|l|l|l|l}
\toprule
 & Disease $\delta=0$ & Airport $\delta=1$ & Pubmed $\delta=3.5$ & Citeseer $\delta=5$ & Cora $\delta=11$\\
 \toprule
$\mathbb{R}^6$ & 88.2 ± 3.1 & 61.7 ± 2.2 & 77.3 ± 1.5 & 64.7 ± 2.3 & 78.1 ± 1.7 \\
$\mathbb{H}^6$ & \textbf{96.9} ± 0.6 & 68.6 ± 1.4 & 76.9 ± 0.5 & 64.0 ± 3.1 & 78.2 ± 1.3 \\
$\spd_3$ & 95.9 ± 2.1 & \textbf{71.2} ± 2.3 & \textbf{78.0} ± 0.6 & \textbf{69.9} ± 0.8 & \textbf{79.7} ± 0.9 \\
$\mathbb{H}^3 \times \mathbb{R}^3$ & 95.9 ± 0.9 & 68.1 ± 4.7 & 74.7 ± 1.9 & 62.2 ± 1.7 & 71.8 ± 4.6 \\ 
\bottomrule
\end{tabular}
    \label{tab:product-space}
\end{table}

\end{document}